\documentclass[lettersize,journal]{IEEEtran}
\usepackage{amsmath,amsfonts}
\usepackage{algorithm}
\usepackage{array}
\usepackage[caption=false,font=normalsize,labelfont=sf,textfont=sf]{subfig}
\usepackage{textcomp}
\usepackage{stfloats}
\usepackage{url}
\usepackage{verbatim}
\usepackage{graphicx}
\usepackage{cite}
\usepackage{colortbl}
\usepackage{makecell}
\usepackage{booktabs} 
\usepackage{multirow} 
\usepackage{graphicx}
\usepackage{siunitx} 
\usepackage{bm}
\usepackage{array}
\usepackage{adjustbox}
\usepackage{algpseudocode}
\usepackage{amssymb}

\usepackage{xspace}
\usepackage{hyperref}
\usepackage{cleveref}
\newcommand{\etal}{\textit{et al.}}
\newcommand{\scenario}[1]{{\small \sf#1}\xspace}

\crefname{equation}{Eq.}{Eqs.}
\Crefname{equation}{Equation}{Equations}
\begin{document}

\title{A Geometric Framework for Absolute Pose and Velocity Estimation with Event Cameras}

\author{Zibin Liu\IEEEauthorrefmark{1}, Shunkun Liang\IEEEauthorrefmark{1}, Banglei Guan,~\IEEEmembership{Member,~IEEE}, Yang Shang, Qifeng Yu, Ji Zhao

	\thanks{\IEEEauthorblockA{\IEEEauthorrefmark{1}These authors contributed equally to this work.}
		
		Manuscript received *, 2021; revised *, 2021. This work was supported by the National Natural Science Foundation of China under Grant 12372189 and the Science and Technology Innovation Program of Hunan Province under Grant 2025RC1045.
        \textit{(Corresponding author: Banglei Guan)}.
        
         Zibin Liu, Shunkun Liang, Banglei Guan, Yang Shang, and Qifeng Yu are with the College of Aerospace Science and Engineering, National University of Defense Technology, Changsha 410073, China. Ji Zhao is an independent researcher, Beijing 100000, China. (e-mail: liuzibin@nudt.edu.cn; liangshunkun21@nudt.edu.cn; guanbanglei12@nudt.edu.cn; shangyang1977@nudt.edu.cn; yuqifeng@nudt.edu.cn; zhaoji84@gmail.com)
}
}

\maketitle

\begin{abstract}
Despite the rapid advancements in event-based motion estimation, current geometric methods primarily focus on velocity estimation. However, absolute pose estimation, which is equally crucial for key applications such as robotic navigation and augmented reality, remains relatively underexplored. Consequently, the simultaneous recovery of absolute pose and velocity from event streams remains an open and challenging problem. To address this gap, we propose a geometric framework for absolute pose and velocity estimation by leveraging 3D lines in the scene and the events they trigger. At the core of the framework lie two key geometric constraints: the orthogonality between a 3D line and the normal vector of its corresponding event plane, and the collinearity of an event with the 2D projection of its associated line. Based on these constraints, we present both linear and polynomial solvers for absolute pose estimation. The former enables efficient computation, while the latter provides a globally optimal solution for rotation. For velocity estimation, we develop an efficient linear solver and a more accurate optimization-based solver to recover both angular and linear velocities. Notably, our methods require a minimum of three event-line correspondences to determine the 6-DoF absolute pose or velocities independently. Extensive experiments in simulation and on real-world datasets demonstrate that our methods achieve state-of-the-art performance, with significant improvements in accuracy and computational efficiency compared to existing methods. The demo code is publicly available at \href{https://github.com/Zibin6/EventPoseVelocity}{https://github.com/Zibin6/EventPoseVelocity}.
\end{abstract}

\begin{IEEEkeywords}
Event camera, Pose estimation, Velocity estimation, Geometric solution.
\end{IEEEkeywords}

\section{Introduction}
\label{sec:intro}

Pose estimation has long been a fundamental technology in the domains of computer vision and robotics, playing a crucial role in various applications such as Visual Odometry (VO)~\cite{nister2004visual} and Simultaneous Localization and Mapping (SLAM)~\cite{9440682}. This task is broadly categorized into absolute pose estimation~\cite{Kneip2014UPnPAO,2009EPnP}, which recovers the rigid transformation from the world coordinate system to the camera coordinate system using 2D-3D correspondences, and relative pose estimation~\cite{guanTPAMI26a,1288525,guanTPAMI26b}, which computes the motion between two or more camera views via 2D-2D correspondences. Beyond these pose initial estimations, recent advances focus on refining pose accuracy through bundle adjustment~\cite{Weber2023power} and rigorous geometric constraints~\cite{liang2026pose}. While pose estimation has been explored with frame-based camera, accurately capturing fast motions remains challenging due to motion blur and limited temporal resolution. The advent of event cameras presents a new frontier for resolving this limitation.

\begin{figure}[t]
	\centering
	\includegraphics[width=7cm]{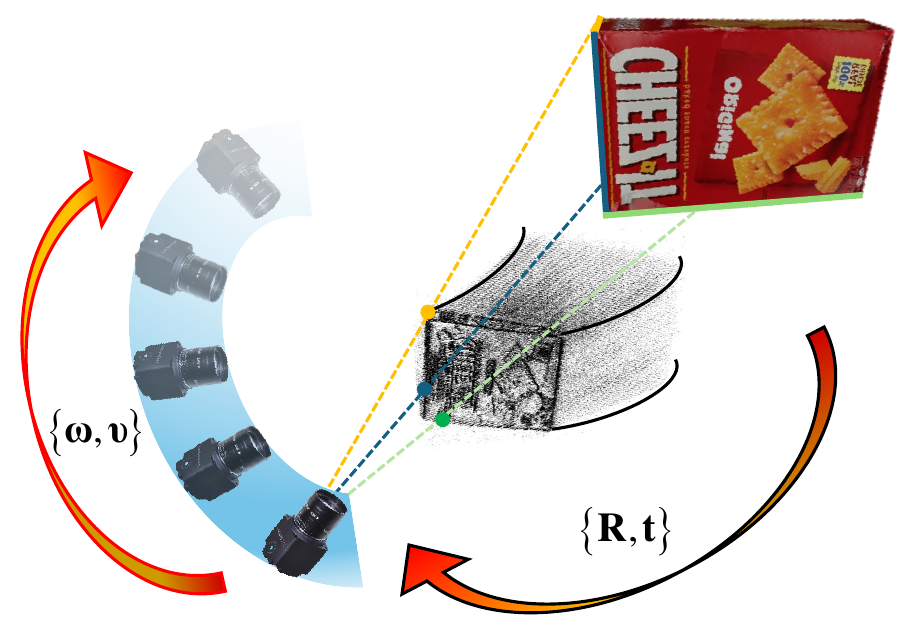}
	\caption{Illustration of joint absolute pose and velocity estimation for event camera. Our methods leverage the geometric constraints between events and lines to recover motion parameters, including both the absolute pose ($\bf{R}, \bf{t}$) and velocities ($\bm{\omega}, \bm{\upsilon}$). 
	}
	\label{fig1}
\end{figure}

\IEEEpubidadjcol

Event cameras, bio-inspired neuromorphic vision sensors, have recently garnered significant attention in the computer vision and robotics communities~\cite{event_survey1,huang20221000}. Unlike conventional frame-based cameras that capture full-intensity images at fixed intervals, event cameras operate with a paradigm-shifting principle of asynchronous triggering. They capture per-pixel brightness changes, generating sparse events that encode the spatio-temporal dynamics of the scene. This unique sensing paradigm offers numerous advantages~\cite{10413505,10684047,10857949}, including high temporal resolution, ultra-high dynamic range, and minimal power consumption. Event cameras are widely recognized as excellent velocity estimators, with their utility well-demonstrated in dynamic tasks such as autonomous driving~\cite{Gehrig2024} and obstacle avoidance~\cite{falanga_davide_dynamic_2020}. However, estimating velocity in the absence of specific positional information holds limited value for many critical applications, such as robotic manipulation~\cite{7833028} and augmented reality~\cite{11028872}. Despite this, event-based absolute pose estimation remains comparatively underexplored. The core challenge lies in effectively associating the sparse, motion-centric event streams with a static 3D world model to anchor the pose globally. Therefore, developing a unified framework that directly exploits event data for joint absolute pose and velocity estimation remains an important and open problem.

A primary challenge in harnessing event data stems from its motion-coupled nature: events are generated only in response to relative motion, and each event carries a precise, asynchronous timestamp~\cite{Gehrig2024,he2024microsaccade}. Existing methods commonly attempt to bridge this gap by compressing events into frame-like representations, which can then be processed by conventional pose estimation algorithms~\cite{event_survey1}. This aggregation, however, sacrifices the high temporal resolution and inherent asynchronicity that are the key advantages of event sensing. This raises a crucial question: \textit{Can a motion estimation framework directly leverage event to jointly recover absolute pose and velocity?}

Another critical issue is the choice of geometric features that are naturally suited to the sparse, edge-like nature of events. While traditional features like corners~\cite{Mueggler17BMVC,glover2021luvharris} have been adapted for event streams, lines offer distinct advantages~\cite{le_gentil_idol_2020-1,valeiras2018event}. They are ubiquitous in structured environments, often more so than feature points~\cite{10124374}, and their structure directly aligns with the high-frequency response of event cameras to edges~\cite{event_survey1}. Recent work has highlighted the effectiveness of lines for event-based relative pose estimation, often by fusing events with angular velocity from an IMU to solve for the remaining linear velocity~\cite{Gao_2023_ICCV,Gao_2024_CVPR}. This demonstrated success motivates our investigation into a more comprehensive problem: \textit{Can event–line correspondences alone suffice to retrieve full 6-DoF absolute pose and velocity without auxiliary sensors?}

Motivated by these observations, we revisit the pose and velocity estimation problem, seeking a purely event-based solution that leverages the unique principles of event cameras. The proposed framework is built on geometric constraints between events and lines to jointly solve for the camera's absolute pose (rotation and translation), and velocity (angular and linear), as illustrated in Fig. \ref{fig1}. We first develop two distinct solvers for absolute pose estimation: a computationally efficient linear solver and a polynomial solver that guarantees globally optimal rotation estimates. We then introduce two methods for velocity estimation: a direct linear solver that prioritizes speed, and a non-linear optimization solver that yields high-accuracy recovery of both angular and linear velocities. Comprehensive evaluations conducted on synthetic data, simulated events, and real-world datasets validate the superior performance of the proposed methods.

The main contributions of this paper are as follows:

\begin{itemize}
	\item  We present a geometric framework for 6-DoF absolute pose and velocity estimation from event-line correspondences, based on orthogonality and collinearity constraints.
	\item We derive two solvers for absolute pose estimation: an efficient linear solver for rapid computation, and a polynomial-based solver with globally optimal rotation estimation.
	\item We derive two solvers for velocity estimation: a fast linear solver and an optimization-based solver for more accurate and robust recovery of angular and linear velocities.
\end{itemize}

The paper is organized as follows. We review related work in Section II. Section III establishes an event-based geometric framework for absolute pose and velocity estimation. In Section IV, we derive two solvers for absolute pose estimation: a linear method and a polynomial method. Section V addresses velocity estimation, presenting both a linear solver and an optimization-based solver. Degenerate cases are discussed and analyzed in Section VI. The performance of the proposed methods is evaluated in Section VII. Finally, Section VIII concludes the paper with a discussion of future research.

\section{Related Work}
\label{sec:related_work}

This section reviews the literature on line-based camera pose estimation, with a focus on both frame-based and event-based methods.

\textbf{Frame-based pose estimation.} The problem of recovering camera pose from 2D-3D line correspondences, known as the Perspective-n-Line (PnL) problem, has been extensively studied for decades. Estimating the full 6-DoF pose requires a minimum of three line correspondences, as each provides two constraints on the pose parameters. This minimal configuration, termed the Perspective-three-Line (P3L) problem, can yield up to eight distinct solutions. Xu \etal \cite{7494617} conducted a systematic analysis of the relationships between various 3D line configurations and solution multiplicities.

For cases involving four or more line correspondences, numerous well-established solutions exist to guarantee a unique solution. These solutions are broadly categorized into direct and iterative approaches. Among direct methods, Hartley \etal \cite{hartley2003multiple} pioneered the Direct Linear Transformation (DLT) algorithm, which recovers the pose from a measurement matrix constructed from a minimum of six lines. Building on this framework, Přibyl \etal \cite{pvribyl2016camera} introduced an enhanced DLT formulation that leverages Plücker line coordinates, albeit requiring at least nine line correspondences. In contrast, iterative methods \cite{christy1999iterative}, while flexible, are prone to converging to local minima and are sensitive to initialization \cite{7494617}. 

\textbf{Event-based pose estimation.} The event-based vision community has developed distinct approaches for pose estimation, which can be classified into relative and absolute pose estimation. In the domain of relative pose estimation, contrast maximization (CMax) is a prominent technique, particularly for estimating rotation. Pioneered by Gallego \etal \cite{Gallego2018AUC,10474186}, CMax operates by maximizing the contrast of motion-compensated event images and has proven effective across various motion scenarios. Various methods have also been proposed for linear velocity estimation. Peng \etal \cite{peng_xin_continuous_2021} developed a closed-form initialization using trifocal tensor geometry, while Gao \etal \cite{Gao_2023_ICCV,Gao_2024_CVPR} proposed minimal and linear solvers that simultaneously estimate velocity and line parameters. Recent work by Lu \etal \cite{lu2024rss} introduced event-based normal flow constraints for real-scale velocity estimation, though these methods typically require IMU support for angular velocity measurements. Without relying on additional sensors, Zhao \etal \cite{zhao2025nf} propose a novel coplanarity relation for the eventail manifold and solve the 5-DoF relative pose estimation problem.

Absolute pose estimation remains central to both VO and SLAM research~\cite{10912788,9879881,8258997}. Existing approaches can be broadly categorized into two main paradigms: filtering-based~\cite{8100099} and optimization-based~\cite{7797445,le_gentil_idol_2020-1} methods. Chamorro \etal \cite{chamorro2022event} combined a Lie-formulated error-state Kalman filter for state prediction with the Levenberg-Marquardt algorithm for precise 6-DoF camera pose optimization. Furthermore, Liu \etal \cite{liu2024line} developed a 6-DoF object pose tracking method that first extracts lines from event streams and then utilizes the Branch-and-Bound algorithm for pose estimation.

Although recent years have witnessed the emergence of geometric solutions for event-based sensors, these methods primarily focus on addressing relative pose estimation challenges \cite{peng_xin_continuous_2021,Gao_2023_ICCV,Gao_2024_CVPR,lu2024rss}. To the best of our knowledge, no closed-form solution exists that directly leverages events and lines for simultaneous absolute pose and velocity estimation. This gap in the literature highlights the need for more efficient approaches in event-based vision, particularly those capable of unifying absolute pose and velocity estimation within a single framework.

\section{Geometry of Event-Based Pose Problem}
\label{sec:methodology}
This section establishes the geometric foundations for our event-based pose and velocity estimation framework. Assuming a calibrated event camera undergoes continuous motion within a structured environment, we leverage 3D lines in the scene and events to determine the camera's absolute pose and velocity. The continuous event stream is segmented into event clusters using fixed-time windows. The absolute pose is estimated using events at the central time instant within each window, while velocity estimation utilizes events across the entire temporal window. Based on single rigid body assumption, the absolute pose is represented by a $SE(3)$ transformation, including a rotation matrix $\mathbf{R}$ and a translation vector $\bf{t}$. Given the discrete and asynchronous nature of events, the estimation of the velocity is transformed into solving the angular velocity $\bm{\omega}$ and the linear velocity $\bm{\upsilon}$. We exploit two fundamental geometric constraints that relate 3D lines to their triggered events: the orthogonality constraint and the collinearity constraint.

\subsection{Absolute Pose Estimation}
Let a 3D line in the world coordinate system be represented by its Plücker coordinates ${\bf{L}} = {[{{\bf{d}}^ \top },{{\bf{m}}^ \top }]^ \top } \in \mathbb{R}{^6}$, where $\mathbf{d}$ is the direction vector of the line, and $\mathbf{m}$ is the moment vector. When observed by the camera, this line generates a spatio-temporal cluster of events. Each event $\mathbf{e}_i$ provides its pixel coordinates ${\left[ {{x_i},{y_i}} \right]^ \top }$, timestamp $t_i$, and polarity $\left\{ { \pm 1} \right\}$. Given the intrinsic parameters $\bf{K}$ of the camera, the normalized image coordinates of event $\mathbf{e}_i$ are computed as ${{\bf{x}}_i} = {{\bf{K}}^{ - 1}}{\left[ {{x_i},{y_i},1} \right]^ \top }$.

\begin{figure}[tbp]
	\centering
	\includegraphics[height=4.5cm]{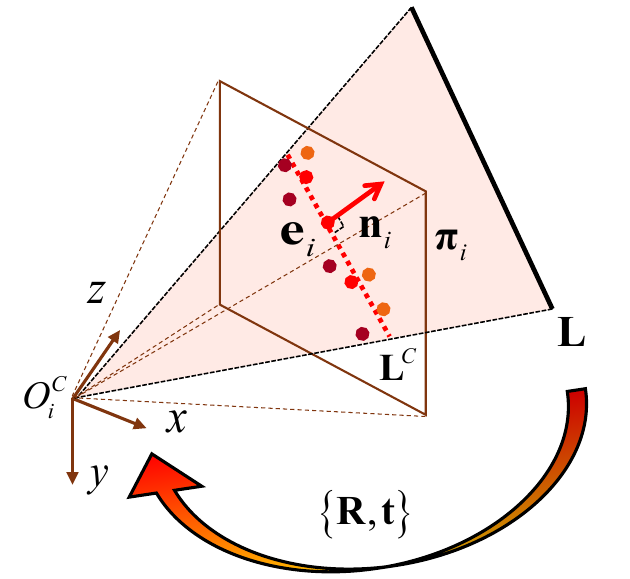}
	\caption{Illustration of the geometric constraints for absolute pose estimation. The absolute pose \{$\bf{R}, \bf{t}$\} maps a 3D line $\bf{L}$ into the camera frame $\cal C$, establishing a geometric relationship with the viewing rays of triggered events.} 
	\label{fig2}
\end{figure}

As depicted in Fig.~\ref{fig2}, the transformation of the 3D line $\bf{L}$ from the world coordinate system to the camera coordinate system $\cal C$ is parameterized by the absolute pose ($\bf{R}, \bf{t}$) at time $t_i$. This transformation is expressed in Plücker coordinates as:
\begin{equation}
	\left[ {\begin{array}{*{20}{c}}
		{{{\bf{d}}^{C}}}\\
		{{{\bf{m}}^{C}}}
\end{array}} \right] = \left[ {\begin{array}{*{20}{c}}
		{\bf{R}}&{\bf{0}}\\
		{{{\left[ \bf{t} \right]}_ \times }{\bf{R}}}&{\bf{R}}
\end{array}} \right]\left[ {\begin{array}{*{20}{c}}
		{\bf{d}}\\
		{\bf{m}}
\end{array}} \right],
	\label{eq1}
\end{equation}
where ${\left[  \cdot  \right]_ \times }$ denotes the skew-symmetric matrix operator, and ${{\bf{L}}^C} = {[{{\bf{d}}^C}^ \top ,{{\bf{m}}^C}^ \top ]^ \top }$ represents the line projected in the camera coordinate system ${\cal C}$ at time $t_i$. As illustrated in Fig.~\ref{fig2}, the projection geometry imposes two fundamental constraints. 

\textbf{Orthogonality constraint.}
As a camera moves through a structured environment, a 3D line $\mathbf{L}$ is projected onto the image sensor, triggering a stream of spatio-temporal events. Let us analyze the geometric information provided by a single event $\mathbf{e}_i$. As shown in Fig.~\ref{fig2}, the normal vector $\mathbf{n}_i$ associated with this event is aligned with the normal vector of the back-projection plane ${\bm{\pi}}_i$. Geometrically, the direction vector of the 3D line in the camera frame, denoted as ${{\bf{d}}^{C}} ={\bf{Rd}}$, must lie on this plane ${\bm{\pi}}_i$. Consequently, ${{\bf{d}}^{C}}$ must be orthogonal to the plane’s normal $\mathbf{n}_i$, leading to the following constraint:
\begin{equation}
	{\bf{n}}_i ^ \top {{\bf{d}}^{C}} = {\bf{n}}_i ^ \top{\bf{Rd}} = 0.
	\label{eq2}
\end{equation}

\textbf{Collinearity constraint.}
In addition to the orthogonality property, we can exploit the point-on-line incidence geometry of the events. Under ideal, noise-free conditions, all events triggered by a 3D line ${\bf{L}}$ must lie precisely on its 2D projection, ${{\bf{L}}^C}$, on the image plane. The 2D projected line ${{\bf{L}}^C}$ in homogeneous coordinates is mathematically given by the normal of the interpretation plane, which is exactly the line's moment vector ${{\bf{m}}^C}$ in the camera frame. Therefore, the homogeneous coordinates ${\bf{x}}_i$ of these events must be orthogonal to ${{\bf{m}}^C}$, yielding the following collinearity constraint:
\begin{equation}
	{\bf{x}}_i^ \top {{\bf{m}}^{C}} = {\bf{x}}_i^ \top \left( {{\bf{Rm}} + {{\left[ {\bf{t}} \right]}_ \times }{\bf{Rd}}} \right) = 0.
	\label{eq3}
\end{equation}

\subsection{Velocity Estimation}
Within a sufficiently short time interval $\left[ {t_i - \Delta t,t_i + \Delta t} \right]$, the camera motion is model by constant angular velocity $\bm{\omega} = [\omega_x, \omega_y, \omega_z]^\top$, and linear velocity $\bm{\upsilon} = [\upsilon_x, \upsilon_y, \upsilon_z]^\top$. Consider an event ${\bf{e}}_j$ triggered at time $t_j$ within this interval, with a relative timestamp $t_{j}^{\prime} = {t_j} - t_i$ with respect to the reference timestamp $t_i$. Under the constant velocity model, the incremental rotation ${{\bf{R}}_j}$ and translation ${{\bf{t}}_j}$ that transform the relative motion of the camera from $t_i$ to $t_j$ are given by:
\begin{equation}
	{{\bf{R}}_j} = \exp \left( {{{\left[ \bm{\omega }  \right]}_ \times }t_j^\prime } \right),{{\bf{t}}_j} = {\bm{\upsilon}}t_j^\prime, 
	\label{eq4}
\end{equation}
where $\exp \left(\cdot\right)$ denotes the exponential map over the $SO(3)$.

\begin{figure}[tbp]
	\centering
	\includegraphics[height=4.5cm]{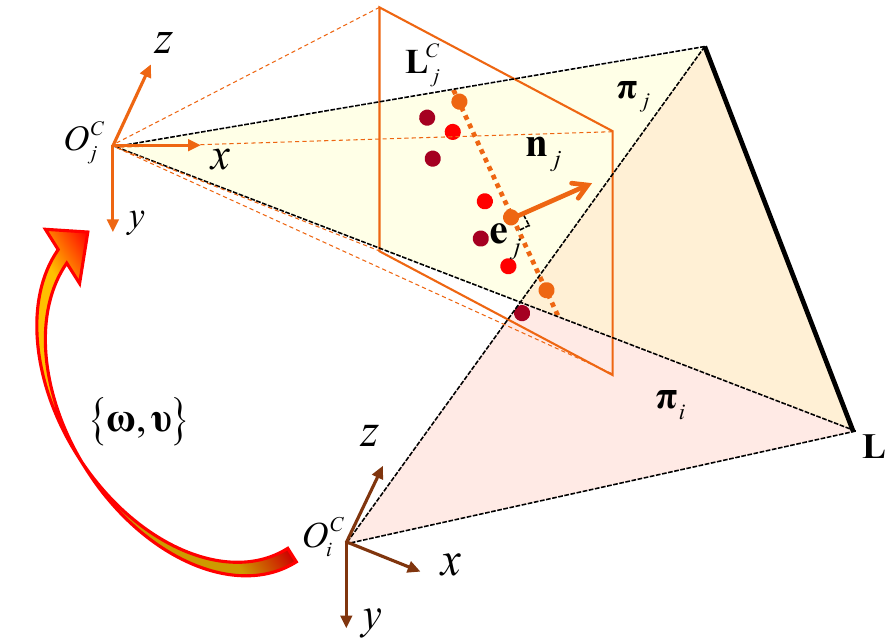}
	\caption{Illustration of velocity estimation. The camera motion between two nearby timestamps $t_i$ and $t_j$ is modeled by a constant angular velocity $\bm{\omega}$ and linear velocity $\bm{\upsilon}$.}
	\label{fig3}
\end{figure}
The same geometric principles must hold for event ${\bf{e}}_j$ at time $t_j$, but they must be applied with respect to the updated camera pose at that instant. As illustrated in Fig.~\ref{fig3}, the pose at time $t_j$ is a composition of the absolute pose ($\mathbf{R}$, $\mathbf{t}$) at time $t_i$ and the incremental motion ($\mathbf{R}_j$, $\mathbf{t}_j$). The line ${\bf{L}}$, now observed from this new pose, is denoted ${{\bf{L}}_j^C}$ in the camera coordinate system and must satisfy the constraints in \cref{eq2,eq3} with respect to event ${\bf{e}}_j$ and its normal vector ${\bf{n}}_j$. This yields:
\begin{equation}
	{\bf{n}}_j^ \top {\bf{d}}_j^C = {\bf{n}}_j^ \top {{\bf{R}}_j}{\bf{Rd}} = 0,
	\label{eq5}
\end{equation}
\begin{equation}
	{\bf{x}}_j^ \top {\bf{m}}_j^C={\bf{x}}_j^ \top \left( {{{\bf{R}}_j}{\bf{Rm}} + {{\left[ {{{\bf{R}}_j}{\bf{t}} + {{\bf{t}}_j}} \right]}_ \times }{{\bf{R}}_j}{\bf{Rd}}} \right) = 0.
	\label{eq6}
\end{equation}

The velocity estimation problem is thus reduced to solving for the unknown motion parameters $\bm{\omega}$ and $\bm{\upsilon}$ that satisfy these equations over the time interval. Our framework seeks to jointly estimate the 6-DoF pose ($\mathbf{R}, \mathbf{t}$) and 6-DoF velocity ($\boldsymbol{\omega}, \boldsymbol{\upsilon}$) using event–line correspondences. Since each event–line correspondence provides two independent scalar constraints, a minimum of three simultaneous correspondences is required to determine the 6-DoF absolute pose under generic non-degenerate configurations. Similarly, three correspondences across distinct timestamps suffice for velocity estimation.

\section{Absolute Pose Estimation Solvers}
Leveraging the orthogonality and collinearity between events and lines, we propose two solvers for absolute pose estimation: a computationally efficient linear solver (\scenario{AbsLin}) and a globally optimal polynomial solver (\scenario{AbsPol}). A crucial component for leveraging the orthogonality constraint is the normal vector ${\bf{n}}_i$ of event ${\bf{e}}_i$, which can be determined through multiple approaches. A straightforward approach computes this vector from two events $\mathbf{e}_1$ and $\mathbf{e}_2$ triggered by the same line. Given their normalized image coordinates $\mathbf{x}_1$ and $\mathbf{x}_2$ respectively, the normal vector can be calculated as ${\bf{n}}_i= \mathbf{x}_1 \times \mathbf{x}_2$. While this method is theoretically sound under ideal conditions, it exhibits sensitivity to noise and outliers in practical scenarios. Therefore, we adopt a more robust approach based on the local motion information encoded in the time surface. First, we compute the normal flow $\bf{f}$ for an event ${\bf{e}}_i$ from the gradient $\nabla \Sigma_e$ of its associated time surface~\cite{ren2024moti}:
 \begin{equation}
	 {\bf{f}} = \frac{{\nabla \Sigma_e }}{{{{\left\| {\nabla \Sigma_e } \right\|}^2}}} \in \mathbb{R}^2.
 \end{equation}

We then determine the direction vector ${{\bf{f}}_ \bot }$ in the image plane that is orthogonal to this normal flow. Inspired by~\cite{zhao2025nf}, we lift the direction ${{\bf{f}}_ \bot }$ into 3D space by appending a zero z-component. The normal vector ${\bf{n}}_i$ is then robustly computed as the cross product:
\begin{equation}
{\bf{n}}_i = {{\bf{x}}_i} \times {{\bf{f}}_ \bot }
\end{equation}
\subsection{Linear solver}
First, we estimate the 6-DoF absolute pose $\{\mathbf{R}, \bf{t}\}$ at time $t_i$ by constructing a linear system from a set of event-line correspondences. To achieve this, we combine the orthogonality and collinearity constraints from \cref{eq2} and \cref{eq3}. The notation $\mathbf{E} = [\bf{t}]_{\times}\mathbf{R}$ denotes the essential matrix. For the purpose of linearization, we first vectorize the essential matrix $\mathbf{E}$ and the rotation matrix $\mathbf{R}$, and then concatenate them to form an 18-dimensional unknown vector $\mathbf{v} = [\operatorname{vec}(\mathbf{E}) ^\top,\operatorname{vec}(\mathbf{R})^\top ]^\top$. Using the Kronecker product ($\otimes$), we rewrite two constraints as a linear system:
\begin{equation}
	\left[ {\begin{array}{*{20}{c}}
			{\bf{h}}_i&{\bf{c}}_i\\
			{\bf{0}}_{1\times9}&{\bf{k}}_i
	\end{array}} \right]{\bf{v}} = 0, 
	\label{eq:ndr}
\end{equation}
where the coefficient matrix are calculated as:
\begin{equation}
	\mathbf{h}_i = (\mathbf{d}^\top \otimes \mathbf{x}_i^\top),\mathbf{c}_i = (\mathbf{m}^\top \otimes \mathbf{x}_i^\top),\mathbf{k}_i = (\mathbf{d}^\top \otimes \mathbf{n}_i^\top).
\end{equation}

We temporarily relax the non-linear geometric constraints, namely the orthogonality of the rotation matrix (i.e., $\mathbf{R}^{\top}\mathbf{R} = \mathbf{I}$ and $\det(\mathbf{R}) = 1$) and the structure of $\mathbf{E}$. This relaxation allows us to treat the 18 unknowns in $\mathbf{v}$ as independent variables. The problem is equivalent to finding the one-dimensional null space of the measurement matrix. For each line, the orthogonality constraint provides one independent constraint equation, while multiple events under the collinearity constraint contribute at most two independent constraint equations. In summary, each line contributes three linearly independent constraint equations under general geometric configurations, thus theoretically requiring at least six independent lines with each line containing at least two events to uniquely determine the solution.

Although the linear solver for $\mathbf{v}$ is inherently ambiguous in scale, this ambiguity can be resolved by re-imposing one of the relaxed geometric constraints. Let the raw solution from the null space be $\mathbf{v}_{\text{raw}}$, from which we can extract a scaled, non-orthonormal rotation matrix $\mathbf{R}_{\text{raw}}$. The true rotation matrix $\mathbf{R}$ is related to $\mathbf{R}_{\text{raw}}$ by an unknown scale factor $s_r$, such that $\mathbf{R} = s_r \cdot \mathbf{R}_{\text{raw}}$. By enforcing the determinant constraint, $\det(\mathbf{R}) = 1$, we can solve for this scale factor:
\begin{equation}
\det(s_r \cdot \mathbf{R}_{\text{raw}}) = {s_r}^3 \det(\mathbf{R}_{\text{raw}}) = 1  \Rightarrow  s_r = \frac{1}{\sqrt[3]{\det(\mathbf{R}_{\text{raw}})}}.
\end{equation}

By applying this unique scaling factor $s_r$ to the entire solution vector $\mathbf{v}_{\text{raw}}$, the true physical scale of all unknown parameters is recovered simultaneously. Finally, to ensure the resulting rotation matrix strictly adheres to its geometric properties, we project the scaled matrix onto the nearest point on the $\mathrm{SO}(3)$ manifold using Singular Value Decomposition (SVD). This final step yields a pose solver that is both geometrically valid and numerically stable.

\subsection{Polynomial solver}
We also propose a nonlinear solver that requires at least three event-line correspondences to estimate the absolute pose, utilizing both orthogonality and collinearity constraints. Specifically, the rotation is first computed using a polynomial solver applied to \cref{eq2}, followed by the determination of translation by solving \cref{eq3} linearly. 

Given $N$ event-line correspondences $(N \geq 3)$, the rotation estimation is formulated as a constrained least-squares problem based on the orthogonality constraint:
\begin{equation}
    \begin{aligned}
        \min_{\mathbf{R}} \quad & \sum_{i} \left( \mathbf{n}_i^\top \mathbf{R} \mathbf{d} \right)^{2} \\
        \text{s.t.} \quad & \mathbf{R}^\top\mathbf{R}=\mathbf{I}, \quad \det(\mathbf{R})=1
    \end{aligned}
    \label{eq:min_nRd}
\end{equation}
To eliminate the non-convex $\mathrm{SO}(3)$ constraints, we adopt the Cayley parameterization of rotation $\mathbf{R}$, which provides a minimal three-parameter representation $(q_x, q_y, q_z)$. Specifically, the rotation matrix can be expressed as:
\begin{equation}
\small
	\setlength{\arraycolsep}{0.5pt}
	\mathbf{R}\!=\!\frac{1}{s_q}\!\left[\!\begin{array}{ccc}
		1+q_x^{2}-q_y^{2}-q_z^{2} & 2(q_x q_y-q_z) & 2(q_x q_z+q_y) \\
		2(q_x q_y+q_z) & 1-q_x^{2}+q_y^{2}-q_z^{2} & 2(q_y q_z-q_x) \\
		2(q_x q_z-q_y) & 2(q_y q_z+q_x) & 1-q_x^{2}-q_y^{2}+q_z^{2}
	\end{array}\right],
	\label{eq:R_Cayley}
\end{equation}
where $s_q = 1 + q_x^2 + q_y^2 + q_z^2$. Then, we reformulate the objective function in \cref{eq:min_nRd} to facilitate the solution process. By extracting the rotation-related components from \cref{eq:ndr} and combining with \cref{eq:R_Cayley}, we obtain:
\begin{equation}
	\mathbf{k}_i \mathbf{r} = 0, \mathbf{r} = [{r_1}, \ldots ,{r_9}]^\top.
	\label{eq:kr0}
\end{equation}

By multiplying both sides by $s_q$, the expression in \cref{eq:kr0} remains invariant under any normalization procedure~\cite{Nakano11,5980272}. Combining \cref{eq:R_Cayley} and \cref{eq:kr0} yields:
\begin{equation}
\setlength{\arraycolsep}{0.1pt}
\small
	\mathbf{a}_i^\top \mathbf{q} = 0,
	\mathbf{q} = [1, q_x, q_y, q_z, q_x^2, q_x q_y, q_x q_z, q_y^2, q_y q_z, q_z^2]^\top.
	\label{eq:ai}
\end{equation}

\begin{table}[tb]
	\centering
	\renewcommand{\arraystretch}{1.5}
	\caption{Elements of vector $\mathbf{a}_i$ in the polynomial solver.}
	\begin{tabular}{|l|l|}
		\hline
		$\mathbf{a}_i^{(1)} = \mathbf{k}_i^{(1)} + \mathbf{k}_i^{(5)} + \mathbf{k}_i^{(9)}$ & $\mathbf{a}_i^{(2)} =  2\mathbf{k}_i^{(8)}-2\mathbf{k}_i^{(6)}$ \\
		\hline
		$\mathbf{a}_i^{(3)} = 2\mathbf{k}_i^{(3)} - 2\mathbf{k}_i^{(7)}$ & 
		$\mathbf{a}_i^{(4)} = 2\mathbf{k}_i^{(4)} - 2\mathbf{k}_i^{(2)}$ \\
		\hline
		$\mathbf{a}_i^{(5)} = \mathbf{k}_i^{(1)} - \mathbf{k}_i^{(5)} - \mathbf{k}_i^{(9)}$ &
		$\mathbf{a}_i^{(6)} = 2\mathbf{k}_i^{(2)}+2\mathbf{k}_i^{(4)}$ \\
		\hline
		$\mathbf{a}_i^{(7)} = 2\mathbf{k}_i^{(3)}+2\mathbf{k}_i^{(7)}$ & 
		$\mathbf{a}_i^{(8)} = \mathbf{k}_i^{(5)}-\mathbf{k}_i^{(1)}-\mathbf{k}_i^{(9)}$ \\
		\hline
		$\mathbf{a}_i^{(9)} = 2\mathbf{k}_i^{(6)}+2\mathbf{k}_i^{(8)}$ & 
		$\mathbf{a}_i^{(10)} = \mathbf{k}_i^{(9)}-\mathbf{k}_i^{(5)}-\mathbf{k}_i^{(1)}$ \\
		\hline
	\end{tabular}
	\label{tab:ai}
\end{table}

The elements of vector $\mathbf{a}_i$ are detailed in Table~\ref{tab:ai}. The optimization problem in \cref{eq:min_nRd} is thus transformed into an unconstrained polynomial minimization:
\begin{equation}
	\min_{q_x,q_y,q_z} \quad G = \sum\limits_{i=1}^{N} \left\|\mathbf{a}_i^\top \mathbf{q}\right\|^2 =
	\mathbf{q}^\top \mathbf{A} \mathbf{q},
\end{equation}
where $\mathbf{A} = \sum_{i=1}^{N} \mathbf{a}_i \mathbf{a}_i^\top$ denotes a $10 \times 10$ moment matrix and $G$ is the non-negative objective function. To find the minima, we apply the first-order optimality condition by setting the partial derivatives of $G$ with respect to the Cayley parameters to zero. This results in a system of three cubic polynomial equations in three variables:
\begin{equation}
	\left\{\begin{aligned}
		\frac{\partial G}{\partial q_x} &= 2\mathbf{q}^\top \mathbf{A} \frac{\partial \mathbf{q}}{\partial q_x} = 0, \\
		\frac{\partial G}{\partial q_y} &= 2\mathbf{q}^\top \mathbf{A} \frac{\partial \mathbf{q}}{\partial q_y} = 0, \\
		\frac{\partial G}{\partial q_z} &= 2\mathbf{q}^\top \mathbf{A} \frac{\partial \mathbf{q}}{\partial q_z} = 0. 
	\end{aligned}\right.
	\label{eq:dEdbcd}
\end{equation}

The polynomial equation system in \cref{eq:dEdbcd} allows us to calculate all stationary points of $G$ in closed-form, among which lie the optimal Cayley parameters. To solve this system, we employ the Gröbner basis method, which reformulates the high-dimensional polynomial equations into a more tractable form. Specifically, we utilize the automatic generator proposed in \cite{Larsson2017GB} to construct the elimination template and action matrix. The polynomial solver yields up to 27 candidate solutions for the Cayley parameters. We first filter these by retaining only the real-valued solutions, which correspond to physically meaningful rotations. The optimal rotation matrix $\mathbf{R}$ is then identified by selecting the candidate that minimizes the geometric error in the objective function \cref{eq:min_nRd}.

Once the optimal rotation matrix $\mathbf{R}$ is determined, the complex 6-DoF pose estimation problem is simplified. At this stage, \cref{eq3} simplifies into a standard linear system of equations solely in terms of the unknown translation vector $\mathbf{t}$, which can be solved for quickly and robustly using a classic linear least-squares method. Crucially, this polynomial-based solver reduces the minimum sample requirement from six for linear methods to only three event-line correspondences.

We briefly summarize the time and space complexity for our two pose solvers. Let $N$ be the number of 3D lines, $M$ the number of events generated per line. The linear solver is a one-shot linear mathod with $O(NM)$ time and space complexity, primarily consumed by the construction and solution of the overdetermined linear system \cref{eq:ndr}. The polynomial solver exhibits a time complexity $O(N) + O(NM)$. Specifically, solving the fixed-degree polynomial system for rotation incurs $O(N)$ complexity, and the dominant cost $O(NM)$ is associated with translation estimation. Its space complexity is likewise $O(NM)$.

\section{Velocity Estimation Solvers}

For velocity estimation, we also introduce two solvers: a highly efficient linear solver (\scenario{VelLin}), and a nonlinear optimization solver (\scenario{VelOpt}).

\subsection{Linear solver}
Once the absolute pose parameters are established, we proceed to estimate the camera’s velocity. We propose a linear, two-stage solver that first determines the angular velocity $\bm{\omega}$ and subsequently computes the linear velocity $\bm{\upsilon}$.

Within sufficiently small temporal intervals, the magnitude of rotational displacement remains correspondingly minimal. Consequently, the relative rotation $\mathbf{R}_{j}$ can be approximated using a first-order Taylor expansion:
\begin{equation}
	{{\bf{R}}_j} \approx {\bf{I}} + {\left[ {\bm{\omega} t_j^\prime } \right]_ \times } = \left[ {\begin{array}{*{20}{c}}
			1&{ - {\omega _z}t_j^\prime }&{{\omega _y}t_j^\prime }\\
			{{\omega _z}t_j^\prime }&1&{ - {\omega _x}t_j^\prime }\\
			{ - {\omega _y}t_j^\prime }&{{\omega _x}t_j^\prime }&1
	\end{array}} \right],
    \label{R_first}
\end{equation}
where ${\bf{I}} \in \mathbb{R}^{3\times3}$ represents the identity matrix. We exploit the geometric constraint from \cref{eq5}, which characterizes the orthogonality between the normal vector of events and the direction vector of lines:
\begin{equation}
	{\bf{n}}_j^ \top {{\bf{R}}_j}{\bf{Rd}} = 0 \Rightarrow \underbrace {\left( {{{\left( {{\bf{Rd}}} \right)}^ \top } \otimes {\bf{n}}_j^ \top } \right)}_{{{\bf{b}}_j}} \cdot {\rm{vec}}\left( {{{\bf{R}}_j}} \right) = 0,
\end{equation}
where ${{\bf{b}}_j} \in \mathbb{R}^{1\times9}$ denotes the vector derived from the Kronecker product operation. By incorporating the approximated rotation matrix into this constraint and performing algebraic manipulation, we obtain a linear system for solving the angular velocity $\bm{\omega}$:
\begin{equation}
	\setlength{\arraycolsep}{1pt}
	\begin{array}{l}
		\left[ {\begin{array}{*{20}{c}}
				{{\bf{b}}_1^{(6)}t_1^\prime  - {\bf{b}}_1^{(8)}t_1^\prime ,{\bf{b}}_1^{(7)}t_1^\prime  - {\bf{b}}_1^{(3)}t_1^\prime ,{\bf{b}}_1^{(2)}t_1^\prime  - {\bf{b}}_1^{(4)}t_1^\prime }\\
				\vdots \\
				{{\bf{b}}_j^{(6)}t_j^\prime  - {\bf{b}}_j^{(8)}t_j^\prime ,{\bf{b}}_j^{(7)}t_j^\prime  - {\bf{b}}_j^{(3)}t_j^\prime ,{\bf{b}}_j^{(2)}t_j^\prime  - {\bf{b}}_j^{(4)}t_j^\prime }
		\end{array}} \right]\left[ {\begin{array}{*{20}{c}}
				\omega_x\\
				\omega_y\\
				\omega_z
		\end{array}} \right]\\[20pt]
		=  - \left[ {\begin{array}{*{20}{c}}
				{{\bf{b}}_1^{(1)} + {\bf{b}}_1^{(5)} + {\bf{b}}_1^{(9)}}\\
				\vdots \\
				{{\bf{b}}_j^{(1)} + {\bf{b}}_j^{(5)} + {\bf{b}}_j^{(9)}}
		\end{array}} \right],
	\end{array}
	\label{eq16}
\end{equation}
where ${\bf{b}}_j^{(h)}$ represents the $h$-th element of the vector ${{\bf{b}}_j}$. Each row in this linear system represents the geometric constraint imposed by a single event-line correspondence. The angular velocity $\bm{\omega}$ contains three unknown components. Thus, we can uniquely determine $\bm{\omega}$ using a minimum of three correspondences via standard linear least squares.

With the angular velocity $\bm{\omega}$ known, the relative rotation ${{\bf{R}}_j}$ is also determined. We then substitute these known values into the collinearity constraint, which is further rearranged to yield:
\begin{align}
	\mathbf{x}_j^\top \big( \underbrace{\mathbf{R}_j \mathbf{R} \mathbf{m}}_{\bm{\alpha}_j} + [ \underbrace{\mathbf{R}_j \mathbf{t}}_{\bm{\beta}_j} +  \mathbf{t}_j ]_\times \underbrace{\mathbf{R}_j \mathbf{R} \mathbf{d}}_{\bm{\gamma}_j} \big) = 0.
	\label{eq17}
\end{align}

To derive a computationally tractable form, we perform algebraic manipulation of \cref{R_first,eq17} by expanding the cross product terms and collecting coefficients. By leveraging the collinearity constraint inherent in the geometric formulation, we systematically reorganize the algebraic terms. Each event-line correspondence can thus be expressed as a linear equation in terms of the velocity components, yielding the following structured form:

\begin{equation}
	\begin{aligned}
		&\left[ 
		\begin{array}{c}
			\gamma_j^{(2)} t_j' x_j^{(3)} - \gamma_j^{(3)} t_j' x_j^{(2)} \\
			\gamma_j^{(3)} t_j' x_j^{(1)} - \gamma_j^{(1)} t_j' x_j^{(3)} \\
			\gamma_j^{(1)} t_j' x_j^{(2)} - \gamma_j^{(2)} t_j' x_j^{(1)}
		\end{array} 
		\right]^\top 
		\left[ 
		\begin{matrix}
			v_x \\
			v_y \\
			v_z 
		\end{matrix} 
		\right] \\
		& = - \Big( 
		x_j^{(1)} \big( \alpha_j^{(1)} + \beta_j^{(2)} \gamma_j^{(3)} - \beta_j^{(3)} \gamma_j^{(2)} \big) \\
		&\qquad + x_j^{(2)} \big( \alpha_j^{(2)} - \beta_j^{(1)} \gamma_j^{(3)} + \beta_j^{(3)} \gamma_j^{(1)} \big) \\
		&\qquad + x_j^{(3)} \big( \alpha_j^{(3)} + \beta_j^{(1)} \gamma_j^{(2)} - \beta_j^{(2)} \gamma_j^{(1)} \big)
		\Big).
	\end{aligned}
\end{equation}
where $\alpha_j^{(h)}$, $\beta_j^{(h)}$, and $\gamma_j^{(h)}$ represent the $h$-th components of vectors $\boldsymbol{\alpha}_j$, $\boldsymbol{\beta}_j$, and $\boldsymbol{\gamma}_j$, respectively, as defined in \cref{eq17}.

Linear velocity $\bm{\upsilon}$ can be solved by constructing a system of linear equations, where each event-line correspondence contributes one equation. With three or more events occurring at different moments, the system becomes over-determined, enabling the use of a linear least squares solver for a robust velocity estimation.

The proposed two-stage linear solver, which decouples the estimation of angular and linear velocities, offers several key advantages in practice. Primarily, it yields a closed-form solution by decomposing the problem into well-conditioned subproblems, thereby significantly reducing computational complexity. Furthermore, the incorporation of multiple event-line correspondences inherently enhances the algorithm's robustness to measurement noise and potential outliers.

\subsection{Optimization solver}
To jointly estimate both angular and linear velocities for higher accuracy, we formulate a nonlinear optimization problem. The optimization parameters are encoded in a six-dimensional vector $\mathbf{s}= [\omega_x, \omega_y, \omega_z,\upsilon_x, \upsilon_y, \upsilon_z]^\top$, and the problem is formulated by minimizing the sum of squared collinearity errors:
\begin{equation}
	\begin{array}{*{20}{c}}
		{{{\bf{s}}^*} = \mathop {\arg \min }\limits_{\bf{s}} F},\\
		{F = {{\sum\limits_{l} {\sum\limits_{j} {\left( {{\bf{x}}_j^ \top \left( {{{\bf{R}}_j}{\bf{R}}{{\bf{m}}_l} + {{\left[ {{{\bf{R}}_j}{\bf{t}} + {{\bf{t}}_j}} \right]}_ \times }{{\bf{R}}_j}{\bf{R}}{{\bf{d}}_l}} \right)} \right)} } }^2}}.
	\end{array}
	\label{eq8}
\end{equation}

Here, the event-line pair $\left\{\left( {j,l} \right)\right\}$ denotes the association between event ${\bf{e}}_j$ and 3D line ${\bf{L}}_{l}$. We solve the non-linear least-squares problem in \cref{eq8} using the Levenberg-Marquardt (LM) algorithm. The update rule is given by:
\begin{equation}
	\mathbf{s}_{k+1} = \mathbf{s}_k - (\mathbf{J}^\top \mathbf{J} + \lambda \mathbf{I})^{-1} \mathbf{J}^\top \mathbf{r},
	\label{eq9}
\end{equation}
where $\mathbf{r}$ is the residual vector, $\mathbf{J}$ denotes the Jacobian matrix representing the partial derivatives of the residuals $F$ with respect to $\bf{s}$, and $\lambda$ is an adaptive damping parameter. The optimization is initialized either with a zero vector or with the result from our linear solver for faster convergence. The process terminates when the relative change in the cost function falls below a predefined threshold or a maximum number of iterations is reached.

A critical component of the LM algorithm is the efficient and accurate computation of the Jacobian matrix. The techniques for computing the Jacobian matrix in optimization problems include analytical derivation, various forms of numerical differentiation (such as forward, central, and backward differences), and hybrid methods that integrate these approaches. We provide both analytical derivation and finite difference methods for solving.

\paragraph{Analytical derivation}
The analytical Jacobian is derived by computing the partial derivative with respect to each component of the state vector $\mathbf{s}$. In this stage, we use the first-order approximation of the rotation matrix ${{\bf{R}}_j}$. This approach reduces computational complexity and accelerates convergence by effectively linearizing rotations for small angles and time intervals, thereby maintaining computational efficiency while enabling faster iterative solutions. For example, the partial derivatives with respect to the angular velocity component ${\omega _x}$ are

\begin{equation}
	\begin{aligned}
		\dfrac{{\partial F}}{{\partial {\omega _x}}} = \sum\limits_l {\sum\limits_j 2 } {\bf{x}}_j^ \top (&\dfrac{{{{\partial \bf{R}}_j}}}{{\partial {\omega _x}}}{{\bf{Rm}}_l} + \dfrac{ \partial {{{[{{\bf{R}}_j}{\bf{t}} + {{\bf{t}}_j}]}_ \times }}}{{\partial {\omega _x}}}{{\bf{R}}_j}{{\bf{Rd}}_l} \\
		&+ {[{{\bf{R}}_j}{\bf{t}} + {{\bf{t}}_j}]_ \times }\dfrac{{{{\partial \bf{R}}_j}}}{{\partial {\omega _x}}}{{\bf{Rd}}_l}),
	\end{aligned}
\end{equation}

For the linear velocity component ${\upsilon _x}$, the partial derivative takes a more concise form:
\begin{equation}
	\dfrac{{\partial F}}{{\partial {\upsilon _x}}} = \sum\limits_l {\sum\limits_j {2\left( {{\bf{x}}_j^ \top \dfrac{{\partial{{[{{\bf{R}}_j}{\bf{t}} + {{\bf{t}}_j}]}_ \times }}}{{\partial {\upsilon _x}}}{{\bf{R}}_j}{\bf{R}}{{\bf{d}}_l}} \right)} } .
\end{equation}

The complete set of partial derivatives for all components, including the remaining angular and linear velocity terms, follows a similar pattern.

\paragraph{Numerical differentiation} 
Alternatively, the Jacobian can be computed using numerical differentiation. The fundamental principle of numerical differentiation is to approximate the slope of a function at a specified point using finite differences. By evaluating the function at neighboring points, we can construct an estimate of the derivative. We use the central differences method, which constructs a symmetric structure by using the values of the function at equidistant points on either side of the current point:
 \begin{equation}
	 \frac{\partial F}{\partial s_g} \approx \frac{F(\mathbf{s} + \epsilon\mathbf{u}_g) - F(\mathbf{s} - \epsilon\mathbf{u}_g)}{2\epsilon},
	 \end{equation}
where $\epsilon = 10^{-8}$ is a small perturbation, and $\mathbf{u}_g$ is the $g$-th standard basis vector in $\mathbb{R}^6$, i.e., a vector with 1 in the $g$-th position and 0 elsewhere. To implement this, we compute numerical derivatives by perturbing each parameter $s_g$ with $\epsilon$ in both positive and negative directions, evaluating the cost function $F$ at these perturbed states, and approximating the partial derivatives through central differences. This numerical differentiation process is performed for all parameters $\bf{s}$ to construct the complete Jacobian matrix.

The linear solver exhibits a time complexity of $O(KN) + O(NKM)$. Solving for angular velocity $\boldsymbol{\omega}$ is $O(KN)$, while solving for linear velocity $\boldsymbol{\upsilon}$ requires iterating over all $NKM$ observed events, leading to $O(NKM)$. The space complexity is $O(NKM)$, primarily for storing the measurement matrix of size $NKM \times 3$ used to solve for $\boldsymbol{\upsilon}$. The per-iteration time complexity of optimization-based solver is $O(NKM)$. The space complexity is $O(NKM)$, similar to linear solver, as it relates to the dimensions of the residual vector and Jacobian matrix, which scale with the total number of events $NKM$.

The pseudocode in \Cref{alg1} explicitly describes the workflow. Starting with an input event stream, the algorithm initially selects an event cluster to form a time surface. Subsequently, it calculates the normal vector and proceeds with event line matching, absolute pose estimation, and velocity estimation. Event line matching is achieved through the LOPET algorithm~\cite{liu2024line}. The estimated pose and velocity parameters serve as initial values for the next event cluster, enabling iterative processing of subsequent clusters.

\begin{algorithm}[t]
\caption{\textbf{Geometric Framework for Event-based Pose and Velocity Estimation}}
\textbf{Input:} Event stream, 3D lines $\{\mathbf{L}_j\}$ \\
\textbf{Output:} Absolute pose $(\mathbf{R}, \mathbf{t})$, velocity $(\boldsymbol{\omega}, \boldsymbol{\upsilon})$
\begin{algorithmic}[1]

\Statex \textcolor{blue}{/* Phase 1: Event Processing and Matching */}
\State Select an event window $\{\mathbf{e}_i\}$ and aggregate events to form time surfaces
\State Compute normal vectors $\{\mathbf{n}_i\}$ from the time surfaces
\State Establish event--line correspondences $\{(\mathbf{e}_i, \mathbf{n}_i, \mathbf{L}_j)\}$
\Statex \textcolor{blue}{/* Phase 2: Absolute Pose Estimation */}
\State \textbf{Linear solver:}
\Statex \hspace{1em} $(\mathbf{R}, \mathbf{t}) \leftarrow$ \scenario{AbsLin}$\big(\{\mathbf{e}_i\},\, \{\mathbf{n}_i\},\, \{\mathbf{L}_j\}\big)$
\State \textbf{Polynomial solver:}
\Statex \hspace{1em} $(\mathbf{R}, \mathbf{t}) \leftarrow$ \scenario{AbsPol}$\big(\{\mathbf{e}_i\},\, \{\mathbf{n}_i\},\, \{\mathbf{L}_j\}\big)$
\Statex \textcolor{blue}{/* Phase 3: Velocity Estimation */}
\State \textbf{Linear solver:}
\Statex \hspace{1em} $(\boldsymbol{\omega}, \boldsymbol{\upsilon}) \leftarrow$ \scenario{VelLin}$\big(\{\mathbf{e}_i\},\, \{\mathbf{n}_i\},\, \{\mathbf{L}_j\},\, \mathbf{R},\, \mathbf{t}\big)$
\State \textbf{Optimization solver:}
\Statex \hspace{1em} $(\boldsymbol{\omega}, \boldsymbol{\upsilon}) \leftarrow$ \scenario{VelOpt}$\big(\{\mathbf{e}_i\},\, \{\mathbf{L}_j\},\, \mathbf{R},\, \mathbf{t}\big)$
\State \Return $\big(\mathbf{R},\, \mathbf{t},\, \boldsymbol{\omega},\, \boldsymbol{\upsilon}\big)$

\end{algorithmic}
\label{alg1}
\end{algorithm}

\section{Analysis of Degeneracies}

\begin{table}[t]
	\centering
	\resizebox{0.5\textwidth}{!}{
		\setlength{\tabcolsep}{4pt}
		\renewcommand{\arraystretch}{1.3} 
		\begin{tabular}{cl|cccc}
			\hline
			\multicolumn{2}{c|}{} & \scenario{AbsLin} & \scenario{AbsPol} & \scenario{VelLin} & \scenario{VelOpt} \\
			\hline
			\multirow{2}{*}{\thead{3D Line\\Distribution}} & Coplanar Lines & $\circ$ & $\checkmark$ & $\checkmark$ & $\checkmark$ \\
			& Parallel Lines & $\times$ & $\times$ & $\times$ & $\times$  \\
			\hline
			\multirow{3}{*}{\thead{Camera\\Motion}} & Pure Rotation & \textit{N/A} &\textit{N/A}  & $\checkmark$ & $\checkmark$ \\
			& Pure Translation & \textit{N/A} & \textit{N/A} & $\checkmark$ & $\checkmark$ \\
            & Short baseline& \textit{N/A} & \textit{N/A} & $\times$ & $\times$ \\
			\hline
		\end{tabular}
	}
	\caption{Degeneracy configuration. $\checkmark$ denotes a robust, non-degenerate case. $\circ$ indicates a degeneracy that our method can resolve through a dedicated strategy. $\times$ signifies an unresolvable degeneracy leading to failure. \textit{N/A} means the scenario is irrelevant to the solver.}
    \label{tab:degeneracy}
\end{table}

This section analyzes potential degenerate configurations that compromise the stability and uniqueness of our proposed pose and velocity estimators. Our findings, summarized in \Cref{tab:degeneracy}, cover degeneracies related to both 3D line configurations and special camera motion cases.

\subsection{3D Line Distribution}
The spatial arrangement of 3D lines is critical for constraining the camera's pose. We investigate two challenging configurations: coplanar lines and parallel lines.

\paragraph{Coplanar Lines}
A critical degeneracy for our linear pose solver \scenario{AbsLin} occurs when all 3D lines are coplanar. This geometric configuration induces a rank deficiency in the system matrix defined by the constraints in \cref{eq:ndr}. Consequently, the solver yields a multi-dimensional solution space rather than a unique pose. While the true pose could, in principle, be recovered by imposing additional non-linear constraints \cite{li2008linear}, this approach adds complexity. Instead, we propose a direct two-step strategy.

First, we proactively detect coplanarity using an efficient algebraic method. We sample two distinct points from each of the $H$ lines to form a $3 \times 2H$ point matrix $\mathbf{B}$. After centering these points, we compute the $3 \times 3$ covariance matrix $\mathbf{S} = \mathbf{B}\mathbf{B}^\top$. The SVD of $\mathbf{S} = \mathbf{U}\Sigma\mathbf{V}^\top$ reveals the principal axes of the point distribution. If the smallest singular value $\Sigma_{min}$ is near zero, e.g., $\Sigma_{min} \leq 10^{-6}$, we determine these lines to be coplanar. The normal to this plane is given by the right singular vector $\mathbf{V}_{min}$ corresponding to $\Sigma_{min}$.

Once coplanarity is confirmed, we resolve the associated rank deficiency by reformulating the problem. We define a corrective rotation $\mathbf{R}_z$ that aligns the plane's normal $\mathbf{V}_{min}$ with the Z-axis, effectively rotating the world such that all lines lie on the $Z=0$ plane. This rotation is computed from the axis-angle vector $\mathbf{r}_z = \mathbf{V}_{min} \times [0,0,1]^\top$. After this transformation, the z-components of all line direction vectors become zero and can be eliminated from the linear system in \cref{eq:ndr}, which resolves the rank deficiency. The pose is then solved in this new frame. Finally, the true pose is recovered by applying the inverse transformation $\mathbf{R}_z$.

\paragraph{Parallel Lines}
When all 3D lines are parallel, they share a common direction vector. This configuration is an unobservable degeneracy for all our pose solvers. The orthogonality and collinearity constraints become insufficient to constrain all 6-DoF of the camera. Specifically, any rotation around the lines' common direction and any translation along it becomes unobservable. The issue with translation is twofold. First, geometrically, translation parallel to $\mathbf{d}$ is ambiguous. Second, and more fundamentally, this specific motion is also unobservable at the sensor level. A camera translating perfectly parallel to 3D lines induces no apparent motion on the image plane. Consequently, an event camera will not trigger enough events for this component of motion. Consequently, this degeneracy is not merely a limitation of the solver, but stems from both geometric indeterminacy and a lack of sensor response, rendering the problem fundamentally unsolvable.

\subsection{Camera Motion}
This section aims to analyze the system degradation under specific camera motion patterns. As this analysis focuses on instantaneous motion and is independent of absolute pose estimation. The evaluation is centered on two velocity solvers.

\paragraph{Pure Rotation}
In the pure rotation scenario ($\boldsymbol{\upsilon}=\mathbf{0}$), our velocity solvers remain fully observable. This robustness is a key advantage of our formulation, as traditional geometric methods (e.g., those relying purely on the epipolar constraint) typically suffer from degeneracy in such case \cite{Laurent2013Direct}. Our method's stability stems from the orthogonality constraint, which directly and independently constrains the angular velocity $\boldsymbol{\omega}$, regardless of the translational component. The collinearity constraint then correctly identifies the linear velocity $\boldsymbol{\upsilon}$ as zero. Therefore, both $\scenario{VelLin}$ and $\scenario{VelOpt}$ do not degenerate under pure rotation.

\paragraph{Pure Translation}
In the case of pure translational motion ($\boldsymbol{\omega}=\mathbf{0}$), the orthogonality constraint remains informative. Instead, it becomes a constraint on $\boldsymbol{\omega}$ which is identified as zero. Meanwhile, the collinearity constraint \cref{eq3} remains effective for recovering the linear velocity $\boldsymbol{\upsilon}$. With $\boldsymbol{\omega}=\mathbf{0}$, the collinearity constraint is simplified as
\begin{equation}
    \mathbf{x}_i^\top(\mathbf{R}\mathbf{m} + [\mathbf{t} + \mathbf{t}_j]_{\times}\mathbf{R}\mathbf{d}) = 0,
    \label{eq:pure_trans_ambiguity}
\end{equation}
This formulation is well-posed and can be solved for the linear velocity $\boldsymbol{\upsilon}$. Consequently, the full motion (with $\boldsymbol{\omega}=\mathbf{0}$) remains observable, and both the $\scenario{VelLin}$ and $\scenario{VelOpt}$ solvers successfully operate in the pure translation case.

\paragraph{Short baseline}
In scenarios with extremely short temporal baselines ($\Delta t \to 0$), the system is partially observable. While the absolute pose ($\mathbf{R}, \mathbf{t}$) remains observable, the estimation of the instantaneous velocity ($\boldsymbol{\omega}, \boldsymbol{\upsilon}$) becomes numerically unstable. This instability occurs because $\Delta t \to 0$ leads to an ill-conditioned linear system for velocity estimation, causing noise in event measurements to be significantly amplified. Therefore, sufficient temporal baselines are necessary for robust velocity estimation.

\begin{figure*}[ht]
	\centering
	\includegraphics[width=0.4\linewidth]{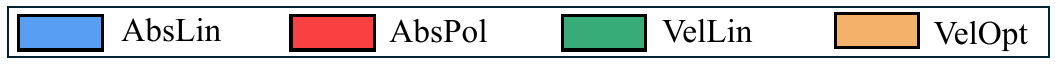}
	\\ 
	\subfloat[Accuracy w.r.t. number of events]{\includegraphics[width=0.83\linewidth]{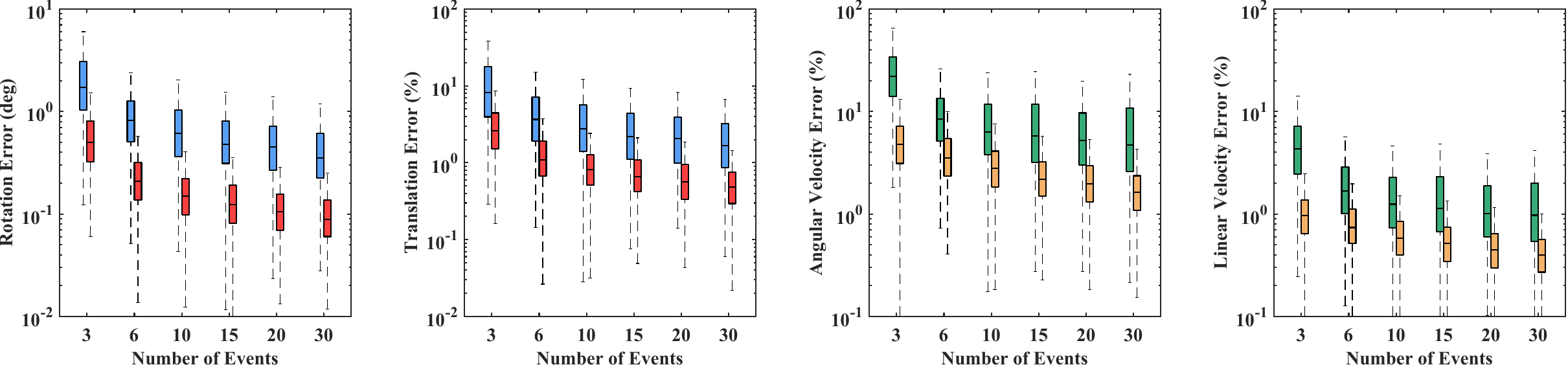} \label{simu_event}}
	\\
	\subfloat[Accuracy w.r.t. number of lines]{\includegraphics[width=0.83\linewidth]{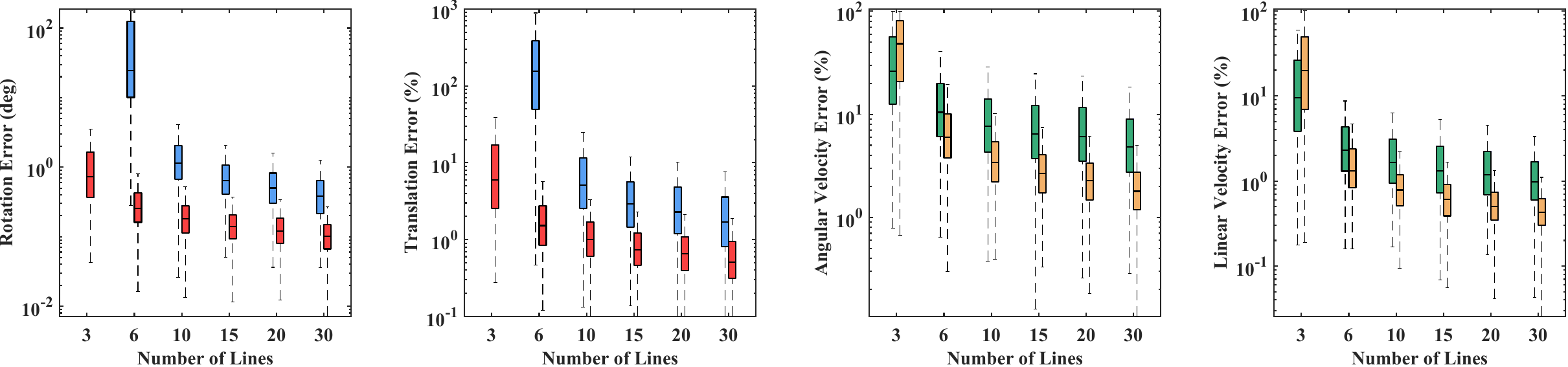} \label{simu_line}}
	\\
	\subfloat[Accuracy w.r.t. the standard deviation of pixel noise]{\includegraphics[width=0.83\linewidth]{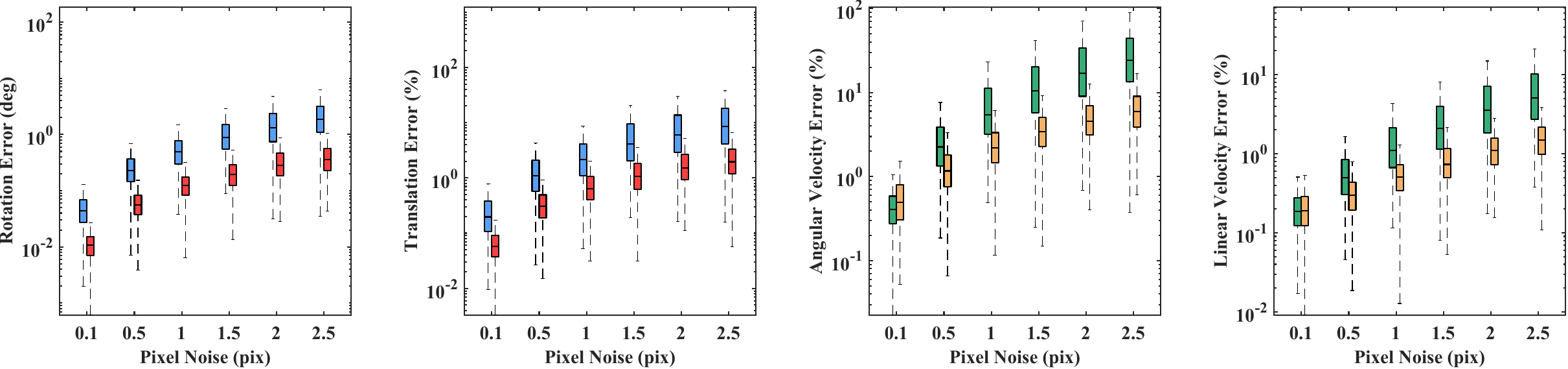} \label{simu_pixel}}
	\\
	\subfloat[Accuracy w.r.t. the standard deviation of time jitter]{\includegraphics[width=0.83\linewidth]{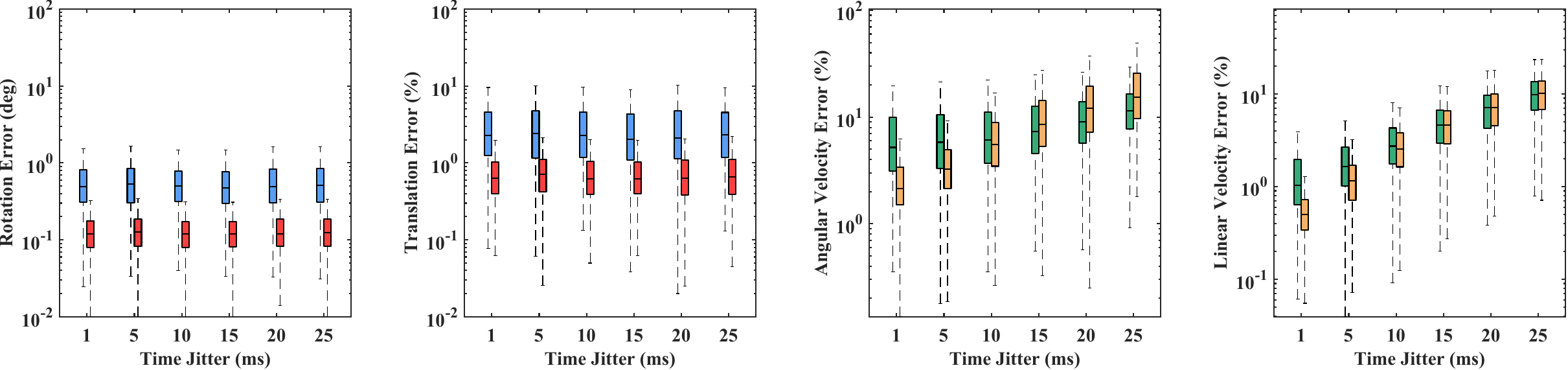} \label{simu_time}}
	\caption{Error analysis of absolute pose and velocity estimation under various conditions. Plots (a)-(d) sequentially illustrate the impact of event count, number of lines, pixel noise, and time jitter on the estimation accuracy.}
	\label{fig_simu}
\end{figure*}

\section{Experiments}
\label{sec:experiments}
Our evaluation framework is comprehensive, incorporating synthetic data, event camera simulations, and real-world experiments on public datasets. All experiments are conducted on a laptop equipped with an Intel Core i7-12700H processor and 16GB of RAM.

\subsection{Synthetic Data} 
We first evaluate the performance of our proposed method across various configurations using synthetic data. The virtual event camera operates at a resolution of 640×480 pixels and a focal length of 800 pixels. In each experimental trial, 3D lines are randomly generated, and events are sampled along these lines within a 100 ms temporal window. The camera's motion is synthesized with random linear velocities in the range of [-5, 5] m/s and angular velocities between [-0.125, 0.125] rad/s. The temporal window is segmented into multiple event sampling intervals, each assigned a random timestamp. The number of sampling intervals is fixed at 20. Within each interval, the camera pose is calculated using linear and angular velocities, and events are formed by projecting multiple randomly positioned points on a 3D line onto the image plane. To assess the robustness of our methods, we introduce two types of noise with varying intensities: pixel noise and timestamp jitter, both following zero-mean Gaussian distributions. A total of 10000 independent trials are performed to ensure statistical significance.

For quantitative evaluation, we employ standard metrics for both absolute pose and velocity estimation. The absolute pose errors are measured using rotation and translation metrics~\cite{liu2024line}: $\varepsilon_{\bf{R}} = \arccos \left( {\left( {\text{trace}\left( {{{\bf{R}}^ \top }{{\bf{R}}_{{\rm{gt}}}}} \right) - 1} \right)/2} \right)$, $\varepsilon_{\bf{t}}=\left\| {\bf{t}-{\bf{t}_{\text{gt}}}} \right\|/\left\| {{\bf{t}_{\text{gt}}}} \right\|\times 100\% $. For velocity estimation, we use relative norm differences for angular and linear velocities in velocity estimation \cite{zhao2025nf}: $\varepsilon_{\bm{\omega}}=\left\| \bm{\omega}-{{\bm{\omega}}_{\text{gt}}} \right\|/(\left\| {{\bm{\omega}}_{\text{gt}}} \right\| + \left\| {{\bm{\omega}}} \right\|)\times 100\%$, $\varepsilon_{\bm{\upsilon}}=\left\| \bm{\upsilon}-{{\bm{\upsilon}}_{\text{gt}}} \right\|/(\left\| {{\bm{\upsilon}}_{\text{gt}}} \right\| + \left\| {\bm{\upsilon}} \right\|)\times 100\%$. 

\begin{table*}[t]
	\centering
	\setlength{\tabcolsep}{8pt}
	\begin{tabular}{c|c|c|c|c|c|c|c}
		\toprule
		\multirow{8}{*}{\makecell[c]{Minimal\\case}}& Method & {\makecell[c]{Number of lines}} &\makecell[c]{$\varepsilon_{\bf{R}}(^\circ)$/$\varepsilon_{\bm{\omega}}$ (\%)} & \makecell[c]{$\varepsilon_{\mathbf{t}}$ (\%)/$\varepsilon_{\bm{\upsilon}}$ (\%)} & SR1 (\%) & SR2 (\%)  & Runtime (ms) \\ \cmidrule(l){1-8}
		&\scenario{AbsLin}&6& $\textbf{0.00}$ & $\mathbf{1.57\times 10^{-11}}$ & $\textbf{100.00}$ & $\textbf{100.00}$ &  $0.24$\\ 
		&\scenario{AbsPol}& 3&$\textbf{0.00}$ & $2.49\times 10^{-8}$ & $99.40$ & $99.70$ &  $0.61$\\ 
		&\scenario{VelLin}  &3 &$6.54\times 10^{-2}$ & $3.22\times 10^{-2}$ & $93.80$ & $97.90$ &  $\textbf{0.08}$\\ 
		&\scenario{VelOpt}&3 &$7.37\times10^{-11}$ & $2.14\times10^{-11}$ & $92.60$ & $92.70$ & $3.71$
		\\ 
		\midrule
		\multirow{4}{*}{\makecell[c]{Non-minimal\\case}}
		&\scenario{AbsLin}&10  &$\textbf{0.00}$ & $2.59\times 10^{-12}$ & $\textbf{100.00}$  & $\textbf{100.00}$  &  $0.33$\\ 
		&\scenario{AbsPol}&10  &$\textbf{0.00}$ & $1.03\times 10^{-9}$ & $\textbf{100.00}$  & $\textbf{100.00}$ &  $0.71$\\
		&\scenario{VelLin} &10  &$2.39\times 10^{-2}$ & $4.38\times 10^{-3}$ & $\textbf{100.00}$  & $\textbf{100.00}$  &  $\textbf{0.42}$\\ 
		&\scenario{VelOpt} &10&$3.52\times 10^{-12}$ & $\mathbf{7.21\times 10^{-13}}$ & $\textbf{100.00}$  & $\textbf{100.00}$ &  $5.69$\\
		\bottomrule
	\end{tabular}
	\caption{Numerical stability and runtime analysis for noise-free synthetic events. SR1 and SR2 represent the success rates (SR) defined by the thresholds of $\varepsilon_{\bf{R}}\left( ^\circ \right),\varepsilon_{\bf{t}}\left( \% \right)$ and $\varepsilon_{\bm{\omega}}\left( \% \right),\varepsilon_{\bm{\upsilon}}\left( \% \right)$, both of which are below ${1^\circ }/\%$ and ${5^\circ }/\%$ , respectively.}
	\label{tab:numerical}
\end{table*}
\textbf{Analysis of the number of events}. To comprehensively evaluate the performance of the proposed methods, this simulation compares the estimation accuracy of pose and velocity under varying configurations of event quantities. Here, the number of events is defined as the number of events generated by each line at each timestamp. The experimental results are shown in Fig.~\ref{fig_simu}\subref{simu_event}. As the number of events increases, the estimation error gradually decreases and stabilizes. In most cases, the error stabilizes when the number of events reaches 6. The results indicate that \scenario{AbsPol} outperforms \scenario{AbsLin}. This may be attributed to the fact that \scenario{AbsPol} has the ability to search for the optimal solution globally, whereas the linear solution is generally more susceptible to noise. For velocity estimation, \scenario{VelOpt} outperforms \scenario{VelLin} as well.

\textbf{Analysis of the number of lines}. Next, we evaluate the impact of the number of lines on the algorithm's accuracy, with experimental results shown in Fig.~\ref{fig_simu}\subref{simu_line}. As the number of lines increases, the estimation error of the algorithm gradually decreases and ultimately stabilizes. Notably, \scenario{AbsPol} demonstrates significantly better performance than \scenario{AbsLin}, with errors converging quickly. Compared to the absolute pose estimation, the velocity estimation approaches demonstrate similar convergence patterns, with \scenario{VelOpt} achieving higher accuracy than \scenario{VelLin} as the line count increases.

\textbf{Pixel noise resilience analysis.} This experiment evaluates the performance of our method against pixel noise of events ranging from 0.1 to 2.5 pixels. The timestamp jitter of events is set to 1 ms. At each noise level, there are 20 lines, with each line generating 15 events. Fig.~\ref{fig_simu}\subref{simu_pixel} presents the errors for absolute pose and velocity estimation. A clear trend of increasing errors is observed as the pixel noises increase, and \scenario{AbsPol} and \scenario{VelOpt} demonstrate significantly lower overall error values. Specifically, \scenario{AbsPol} maintains rotational errors within $0.6^\circ$ and keeps median translation errors under 2.0\%. Similarly, \scenario{VelOpt} demonstrates robust performance, with median angular velocity errors under 6.0\% and median linear velocity errors not exceeding 1.5\%, even at a noise level of 2.5 pixels.

\textbf{Time jitter resilience analysis.} In this experiment, we evaluate the performance of our solvers under different levels of timestamp jitter, with each sampling interval's timestamp perturbed with zero-mean Gaussian noise with a standard deviation ranging from 1 to 25 ms. The event pixel noise is fixed at 1 pixel, while other parameters remain identical to the pixel noise experiment. For absolute pose estimation, our methods demonstrate remarkable stability owing to their inherent resilience to temporal disturbances. Through strategic utilization of events at intermediate timestamps, \scenario{AbsLin} and \scenario{AbsPol} consistently maintain median rotation errors within $0.6^\circ$ and median translation errors below 2.5\% across all jitter levels. While velocity estimation demonstrates increased sensitivity to time jitter, particularly in linear velocity, \scenario{VelLin} and \scenario{VelOpt} maintain satisfactory performance with median errors not exceeding 6.1\% for temporal jitter up to 10 ms.

\begin{table*}[ht]
	\centering
	\setlength{\tabcolsep}{0.4cm}
	\begin{tabular}{l|c|c|c|c|c|c}
		\toprule
		Trajectory & \multicolumn{2}{c|}{Circular} & \multicolumn{2}{c|}{Curved} & \multicolumn{2}{c}{Straight} \\
		\midrule
		Speed & High-speed &Low-speed & High-speed &Low-speed & High-speed &Low-speed \\
		\midrule
		\multicolumn{1}{l|}{Median Error} & \multicolumn{6}{c}{$\varepsilon_{\bf{R}}\left( ^\circ  \right)/\varepsilon_{\bf{t}}\left( \%  \right)$}\\
		\midrule
		\scenario{AbsLin}&  4.32/4.90&3.41/4.15 & 2.83/2.96 & 1.72/2.21 &  2.86/3.10 & 1.73/1.88  \\
		\scenario{AbsPol} &  \textbf{3.29/3.42} &\textbf{3.13/2.75} & \textbf{2.65/2.67} & \textbf{1.57/1.38} &\textbf{2.41/2.53} & \textbf{1.50/1.57}  \\
        \scenario{P3L}~\cite{p3l1989} &  4.96/4.99 & 3.98/4.21 & 3.65/3.61 &2.64/2.66 & 4.68/4.52 & 2.63/2.24  \\
        \scenario{ASpnL}~\cite{7494617} &  4.10/4.16 & 3.82/3.74 & 3.88/3.52 & 2.28/2.31 & 4.73/4.69 & 1.81/1.78  \\
		\midrule
		\multicolumn{1}{l|}{Median Error} & \multicolumn{6}{c}{ $\varepsilon_{\bm{\omega}}\left( \%  \right)/\varepsilon_{\bm{\upsilon}}\left( ^\circ  \right)$}\\
		\midrule
		\scenario{VelLin}&10.55/14.52 &9.32/12.64 &9.72/13.87 &7.45/11.83 & 7.92/6.93&7.08/6.50 \\
		\scenario{VelOpt} &  \textbf{7.74/9.15} &\textbf{6.93/8.96} & \textbf{7.22/12.26} & \textbf{6.72/10.79}& \textbf{3.72/4.95} & \textbf{3.61/3.68} \\
        \scenario{Eventail}~\cite{Gao_2024_CVPR}&  -/21.19 & -/20.06& -/25.60 & -/23.94 & -/15.16 & -/13.64  \\
        \scenario{IncBat}~\cite{zhao2025nf}&  10.15/22.69  & 9.48/21.15 & 22.41/30.54 & 21.22/28.94 & 10.05/18.56 & 9.48/16.43  \\
		\bottomrule
	\end{tabular}
	\caption{Error comparison of absolute pose and velocity estimation on simulated events.}
	\label{tab2error}
\end{table*}

\begin{figure}[t]
	\centering
	\subfloat[Success rate]{\includegraphics[width=0.48\linewidth]{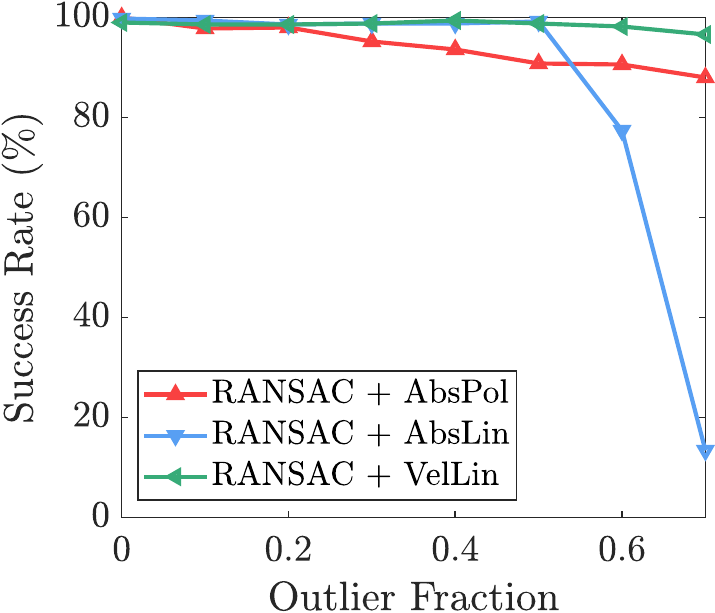}}
	\subfloat[Iteration Count]{\includegraphics[width=0.48\linewidth]{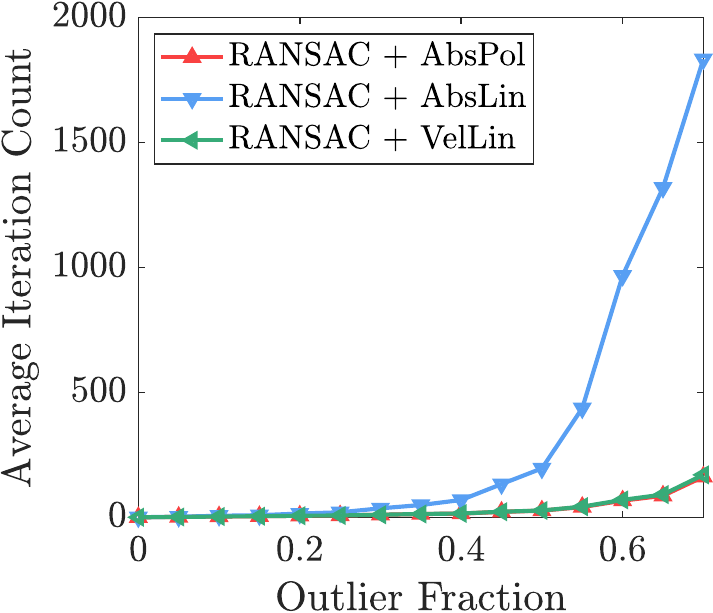}}
	\caption{Robustness analysis to outlier correspondences. We evaluate the minimal solvers \scenario{AbsPol}, \scenario{AbsLin}, and \scenario{VelLin} under varying outlier fractions. (a) Success rate of pose estimation. (b) Total number of RANSAC iterations required.}
	\label{fig:ransac}
\end{figure}

\textbf{Robustness to outlier correspondences.}
To evaluate the tolerance of our solvers to erroneous event–line associations, we inject random mismatches at ratios from 0\% to 60\% into synthetic event–line correspondences, keeping all other settings unchanged. The minimal solvers \scenario{AbsPol}, \scenario{AbsLin}, and \scenario{VelLin} are embedded into a standard RANSAC framework for pose and velocity estimation. Note that \scenario{VelOpt} is excluded as it is designed for refinement after inlier selection. For each outlier ratio, we perform 1000 independent Monte Carlo trials and report the success rate and average number of RANSAC iterations. A trial is deemed successful if the estimated pose satisfies: rotation error $\varepsilon_{\bf{R}} < 5^\circ$, and relative translation and velocity errors $\varepsilon_{\bf{t}}, \varepsilon_{\bm{\omega}}, \varepsilon_{\bm{\upsilon}} < 5\%$. The results are shown in Fig.~\ref{fig:ransac}. The proposed method maintains a success rate above 90\% even when the outlier ratio reaches 50\%, confirming the strong robustness of our solvers when integrated within RANSAC. Moreover, as the outlier ratio increases, the required RANSAC iterations for \scenario{AbsPol} and \scenario{VelLin} remain consistently low, whereas \scenario{AbsLin} exhibits a sharp rise in iteration count. This behavior stems directly from their minimal sample sizes: \scenario{AbsPol} and \scenario{VelLin} require only three event–line correspondences, while \scenario{AbsLin} needs six. Consequently, in high-outlier scenarios commonly encountered in practice, \scenario{AbsPol} emerges as the preferred minimal solver, offering an optimal trade-off between robustness, accuracy, and computational efficiency.

\textbf{Analysis of numerical stability and runtime}. 
Furthermore, we evaluate the numerical stability and runtime of our algorithms in noise-free minimal and non-minimal cases. The minimal case refers to solving with the minimum number of lines required by the algorithm, while the non-minimal case utilizes 10 lines. The experimental results are presented in Table~\ref{tab:numerical}. For absolute pose estimation, both \scenario{AbsLin} and \scenario{AbsPol} demonstrate strong numerical stability. Notably, \scenario{AbsLin} exhibits remarkably low numerical error and achieves a perfect 100\% success rate. While \scenario{AbsPol} is also highly stable, its performance is slightly affected by the precision requirements of its high-degree polynomial solver, causing the success rate to drop marginally to 99.70\% in minimal cases. Regarding velocity estimation, in minimal data scenarios, the performance of \scenario{VelLin} demonstrates superior robustness compared to \scenario{VelOpt}. However, when sufficient redundant observations are introduced, the accuracy of \scenario{VelOpt} improves dramatically, showcasing its superior ability to leverage additional information for enhanced precision. In terms of computational efficiency, the linear solvers (\scenario{AbsLin} and \scenario{VelLin}) consistently outperform \scenario{AbsPol} and \scenario{VelOpt} in processing speed. Nevertheless, all proposed algorithms are efficient enough to satisfy real-time requirements.

\subsection{Event Camera Simulation}

\begin{figure}[t]
	\centering
	\subfloat[RGB image]{\includegraphics[width=0.48\linewidth]{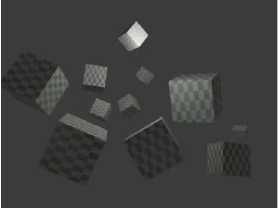}}\label{rgb}
	\subfloat[Event accumulation]{\includegraphics[width=0.48\linewidth]{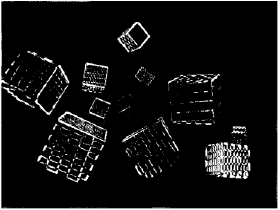}}\label{event}
	\caption{Experimental setup for event simulation. We visualize the data generation process. (a) Rendered RGB image. (b) The corresponding event accumulation image generated from the RGB sequence.}
	\label{fig_rgbevent}
\end{figure}

To validate the accuracy of our method, we conduct comprehensive experiments using a simulated event camera. Our simulation pipeline consists of two main stages. First, we utilize Blender to render high-fidelity video sequences of camera motion, where the virtual camera parameters are configured to match those used in synthetic experiments. The virtual environment includes several randomly placed boxes. This approach enables us to obtain accurate camera motion parameters and trajectories. Subsequently, the rendered video sequences are processed using V2E~\cite{Hu2021v2eFV} to generate realistic event streams, which accurately simulate the event generation process by modeling sensor characteristics. This two-stage simulation framework provides access to accurate ground truth data for quantitative evaluation and enables a systematic analysis of our method’s performance under controlled conditions. The rendered RGB image and event accumulation image are shown in Fig.~\ref{fig_rgbevent}.

Subsequently, we conduct experiments with the camera following diverse motion trajectories, including circular, curved, and straight paths. The experiments are divided into two groups: a low-speed group at approximately 1 m/s and a high-speed group at approximately 5 m/s. This comprehensive evaluation framework enables us to validate the capabilities of our method. For absolute pose estimation, we compare our methods with classic methods including \scenario{P3L}~\cite{p3l1989} and \scenario{ASPnL}~\cite{7494617}. We extract lines from event clusters and feed them into \scenario{P3L} and \scenario{ASPnL} combined with the RANSAC framework for pose estimation. For velocity estimation, we select \scenario{Eventail}~\cite{Gao_2024_CVPR} and \scenario{IncBat}~\cite{zhao2025nf} as baseline methods. Since these two methods solve for scale-ambiguous velocity $\bm{\upsilon}$, we adopt the evaluation metric from~\cite{zhao2025nf} for velocity error assessment. \scenario{Eventail}~\cite{Gao_2024_CVPR} takes ground-truth angular velocity as input and only evaluates the translational velocity error.

The quantitative evaluation of median pose and velocity error statistics is presented in Table~\ref{tab2error}. Both proposed methods demonstrate superior performance compared to baseline approaches. Notably, \scenario{AbsPol} and \scenario{VelOpt} achieve significantly reduced median errors compared to competing methods. These results align with our synthetic data evaluations, offering robust empirical evidence for the efficacy of the proposed methods.

\begin{figure*}[ht]
	\centering
	\includegraphics[width=0.9\linewidth]{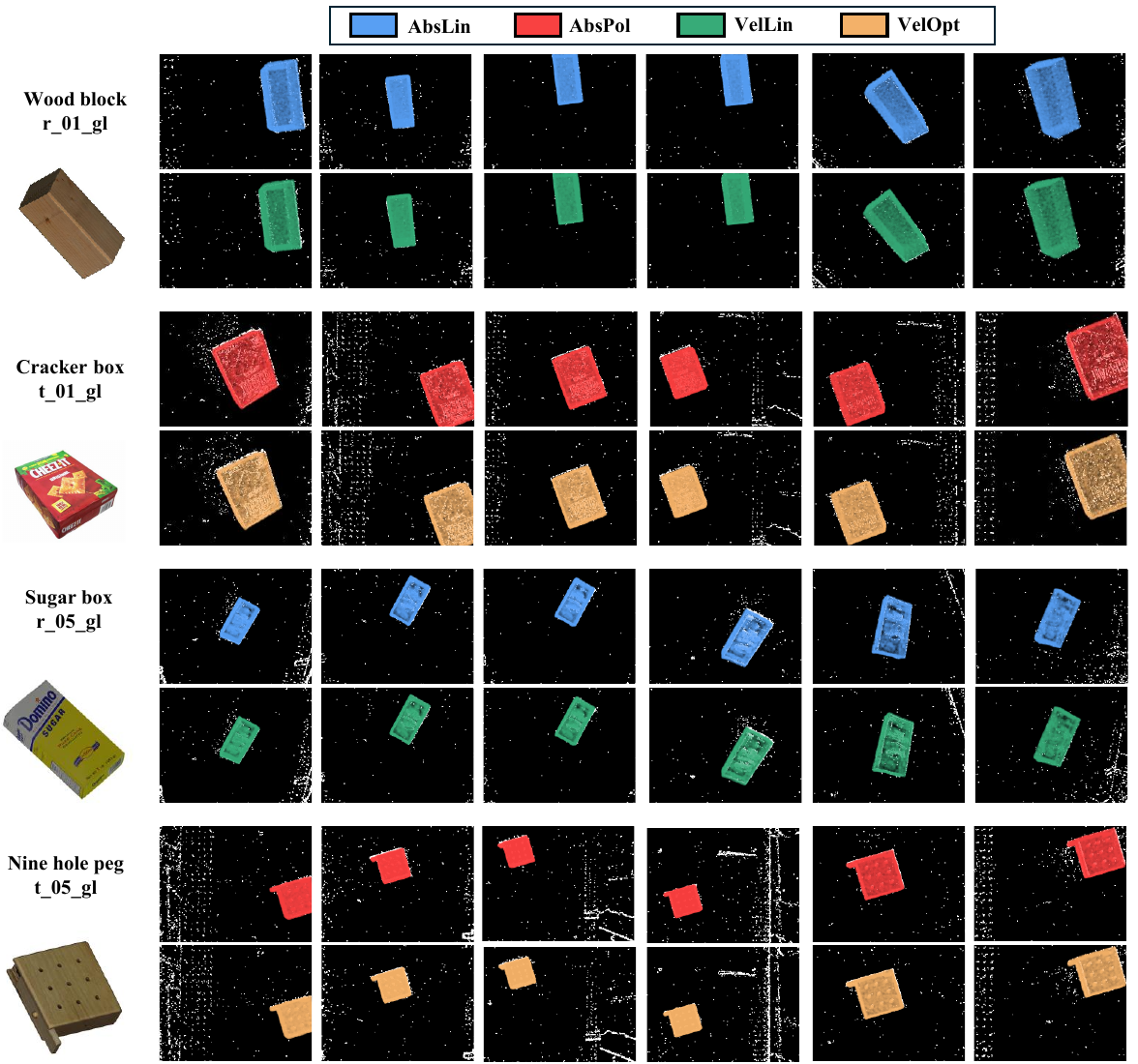}
	\caption{Reprojection results on the E-POSE dataset. Object models are reprojected onto the event accumulation images (for visualization purposes only) using the estimated poses.}
	\label{fig_epose}
\end{figure*}

\begin{table*}[htbp]
	\centering
	\setlength{\tabcolsep}{10pt}
	\begin{tabular}{l|cc|cc|cc|cc}
		\toprule
		Error& \multicolumn{8}{c}{RMSE (unit-less) $\downarrow$ / Median (unit-less) $\downarrow$ }\\
		\midrule
		\multirow{3}{*}{Sequence} & \multicolumn{2}{c|}{Wood block}& \multicolumn{2}{c|}{Cracker box}& \multicolumn{2}{c|}{Sugar box}& \multicolumn{2}{c}{Nine hole peg} \\
		\cmidrule{2-9}
	  	& r\_01\_gl  & t\_01\_gl & r\_01\_gl&t\_01\_gl& r\_05\_gl& t\_05\_gl& r\_05\_gl& t\_05\_gl \\
		\midrule
		\scenario{LS-Opt} & 1.61/1.44 & 1.59/1.36&1.30/1.32 &1.35/1.20 &  2.13/2.07  &  3.56/3.19 & 4.68/4.30 & 3.96/3.49 \\
		\scenario{EL-SLAM}~\cite{chamorro2022event} &0.22/0.20 &0.42/0.33 &0.38/0.35 & 0.24/0.17 & 0.67/0.62   & 1.46/1.22  & 2.04/1.86 &1.97/1.80\\
		\scenario{LOPET}~\cite{liu2024line} & 0.19/0.17& 0.23/0.23&0.22/0.19 & 0.12/0.11&  0.42/0.37  &  1.06/0.82 & 2.62/2.51 &2.46/2.37 \\
		\scenario{AbsLin} & 0.51/0.43&0.43/0.40&0.40/0.37 &0.38/0.32 & 0.50/0.46   & 1.21/1.06  & 1.88/1.60 & 1.61/1.55\\
		\scenario{AbsPol} &0.12/0.12 & \textbf{0.14}/\textbf{0.13} &0.13/0.12 &\textbf{0.11}/\textbf{0.09}& \textbf{0.27}/\textbf{0.22}  & 0.35/0.30  & 1.52/1.49 &1.40/1.35 \\
		\scenario{VelLin} &0.13/0.11  & 0.16/0.14 &0.15/0.14& 0.12/0.12 &  0.34/0.31  & 0.40/0.32  &1.60/1.55&1.37/1.33 \\
		\scenario{VelOpt} &\textbf{0.11}/\textbf{0.10} &\textbf{0.14}/0.14 & \textbf{0.12}/\textbf{0.10} & \textbf{0.11}/0.10 &  0.30/0.28  &  \textbf{0.33}/\textbf{0.27}  & \textbf{1.43}/\textbf{1.41} &\textbf{1.26}/\textbf{1.24}\\
		\bottomrule
	\end{tabular}
	\caption{Absolute Trajectory Error of absolute pose and velocity estimation on the E-POSE dataset.}
	\label{tab_epose}
\end{table*}

\subsection{Real-World Experiment}
To validate the efficacy and robustness of the proposed framework in real-world scenarios, we conduct extensive experiments on two public event-based object pose estimation datasets: E-POSE~\cite{hay2025pose} and LOPET~\cite{liu2024line}. Both datasets provide event streams captured by event cameras alongside high-precision ground truth poses from external motion capture systems, making them ideal for rigorous evaluation.

For a comprehensive evaluation, we implement three line-based approaches as baselines:

\textit{(i)} \scenario{LS-Opt}: A baseline method using least-squares optimization to minimize the distance between reprojected 3D lines and their corresponding events for object tracking.

\textit{(ii)} \scenario{EL-SLAM}~\cite{chamorro2022event}: An event-based line-SLAM method that estimates and tracks the 6-DoF camera pose. As the official code is unavailable, we have reimplemented and adapted its core principles for the task of object pose tracking.

\textit{(iii)} \scenario{LOPET}~\cite{liu2024line}: A recent line-based object pose estimation and tracking method, employing the Branch-and-Bound algorithm with robust optimization.

To quantitatively assess the performance of absolute pose and velocity estimation, we utilize the EVO toolbox~\cite{grupp2017evo} to compute the Absolute Trajectory Error (ATE). Specifically, we report both the Root Mean Square Error (RMSE) and median error for all benchmarked algorithms, enabling a unified and quantitative comparison. Our evaluation pipeline first estimates the absolute pose (\scenario{AbsLin}, \scenario{AbsPol}) upon receiving a new event cluster. For velocity evaluation (\scenario{VelLin}, \scenario{VelOpt}), the poses from our superior \scenario{AbsPol} method are used as input. The resulting estimated velocities are then integrated to produce absolute poses, enabling a direct and fair ATE comparison against all other methods. Event-line association is derived from LOPET~\cite{liu2024line}, with threshold parameters consistent with the original settings.

\begin{figure*}[ht]
	\centering
	\includegraphics[width=0.5\linewidth]{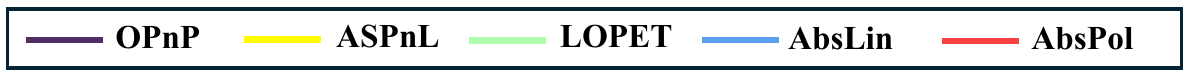}
	\\
	\subfloat[Reprojection Visualization]{\includegraphics[width=0.45\linewidth]{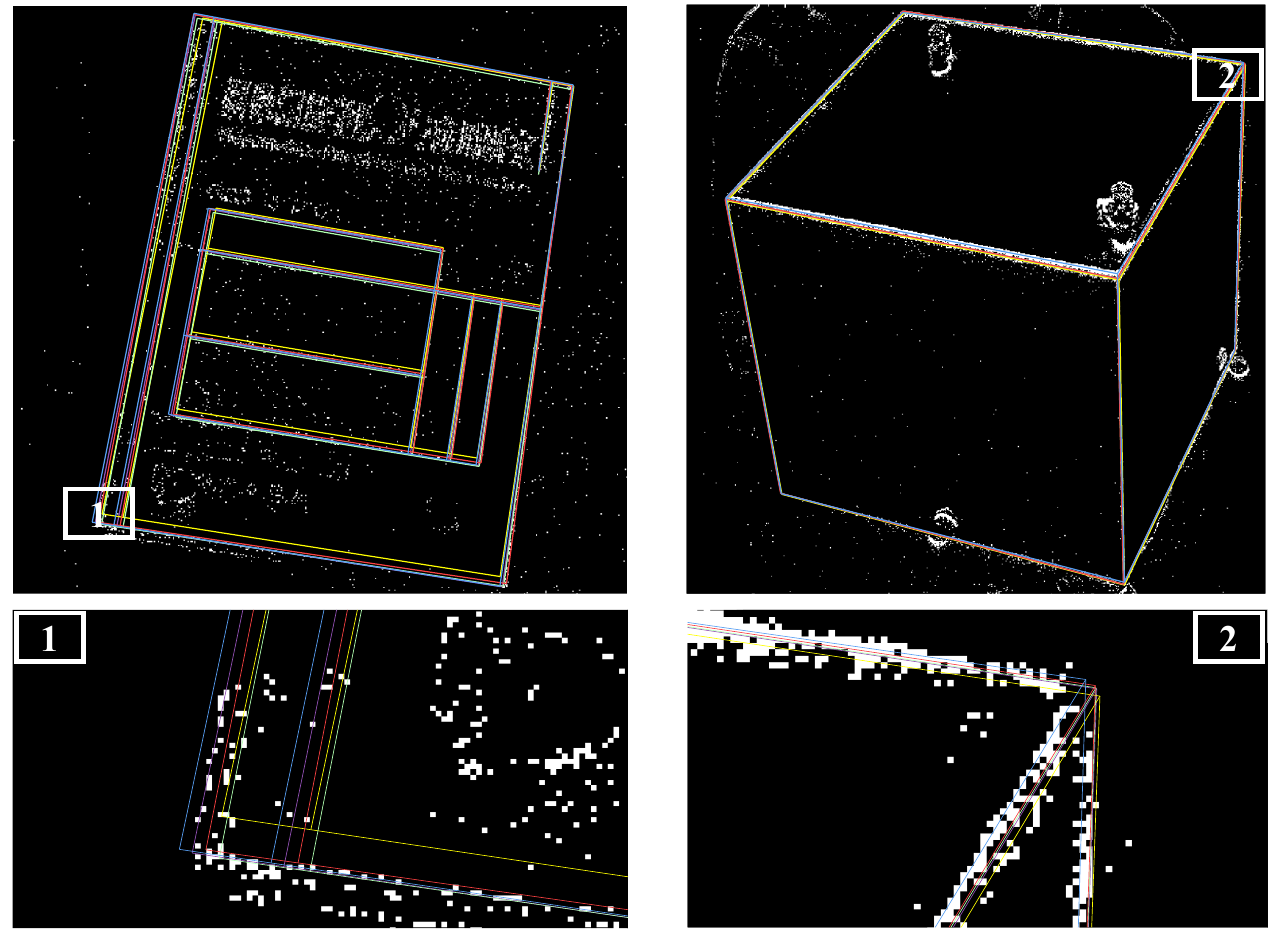} \label{re_vis}}
	\qquad
	\subfloat[Reprojection Error]{\includegraphics[width=0.4\linewidth]{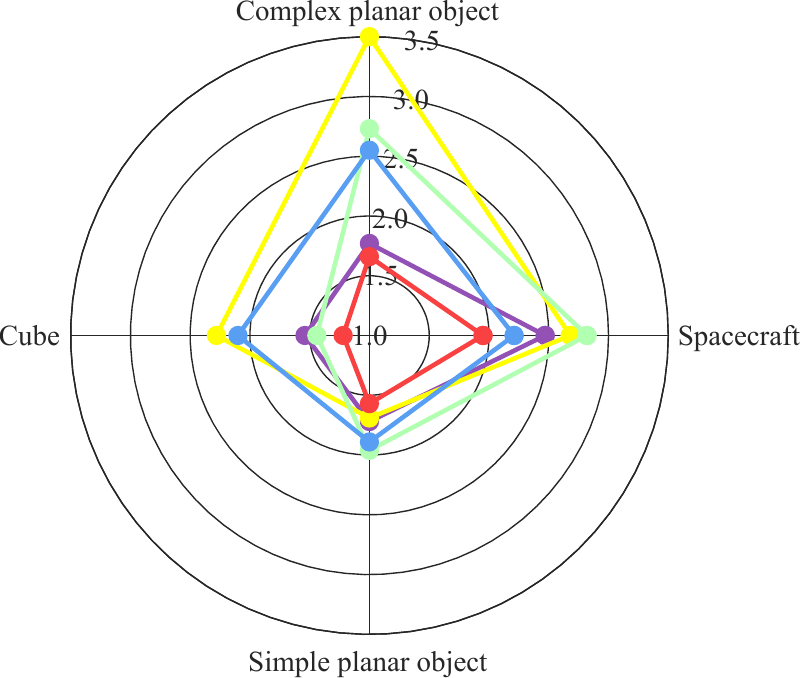} \label{re_qua}}
	\caption{Initial pose estimation results on the LOPET dataset. (a) Visual comparison of reprojected 3D lines from different methods. (b) Mean reprojection error of line endpoints (unit: pixels).}
	\label{fig_initial}
\end{figure*}

\paragraph{Experimental results on E-POSE}

We evaluate our algorithm on the E-POSE dataset, selecting four representative objects characterized by distinct lines: Wood block, Cracker box, Sugar box, and Nine hole peg. The test sequences for each object encompass two typical motion patterns: rotation-dominant and translation-dominant.

Table~\ref{tab_epose} presents the ATE results for rotation and translation on representative sequences. Our proposed methods, \scenario{AbsPol} and \scenario{VelOpt}, consistently outperform all baselines. Notably, VelOpt achieves the best overall performance. Fig.~\ref{fig_epose} provides a visual testament to our method’s accuracy by showing the 3D object models reprojected onto event accumulation images using poses estimated by our methods. The tight and consistent alignment between the projected models and the high-contrast edges in event accumulation images visually corroborates the high accuracy reported in Table~\ref{tab_epose}, even during rapid movements and changes in orientation.

\begin{table*}[htbp]
	\centering
	\setlength{\tabcolsep}{8pt}
	\begin{tabular}{l|cccc}
		\toprule
		Error& \multicolumn{4}{c}{RMSE (unit-less) $\downarrow$ / Median (unit-less) $\downarrow$ } \\
		\midrule
		Sequence  &{Simple planar object}&{Complex planar object}& Cube& Spacecraft \\
		\midrule
		\scenario{LS-Opt} &  3.62/2.00  & 3.55/3.51 & 14.98/4.52& 2.26/2.23\\
		\scenario{EL-SLAM}~\cite{chamorro2022event} & 3.45/2.03& 3.50/3.45& 6.62/1.77& 1.47/1.42 \\
		\scenario{LOPET}~\cite{liu2024line} & 0.59/0.53& 2.18/2.00&1.12/0.86& 1.36/1.31 \\
		\scenario{AbsLin} &  0.61/0.55 & 2.16/2.07&1.35/1.25& 1.65/1.63\\
		\scenario{AbsPol} &0.43/0.42  & \textbf{2.08}/\textbf{1.93} & 1.08/\textbf{0.99} & \textbf{1.21/1.19}\\
		\scenario{VelLin} & 0.54/0.51&  2.24/2.19 &1.35/1.33& 1.30/1.28\\
		\scenario{VelOpt} &\textbf{0.40}/\textbf{0.38}& 2.20/2.15&\textbf{1.05}/1.03& 1.26/1.25 \\
		\bottomrule
	\end{tabular}
	\caption{Absolute Trajectory Error of absolute pose and velocity estimation on the LOPET dataset.}
	\label{tab_lopet}
\end{table*}

\begin{figure*}[ht]
	\centering
	\includegraphics[width=0.85\linewidth]{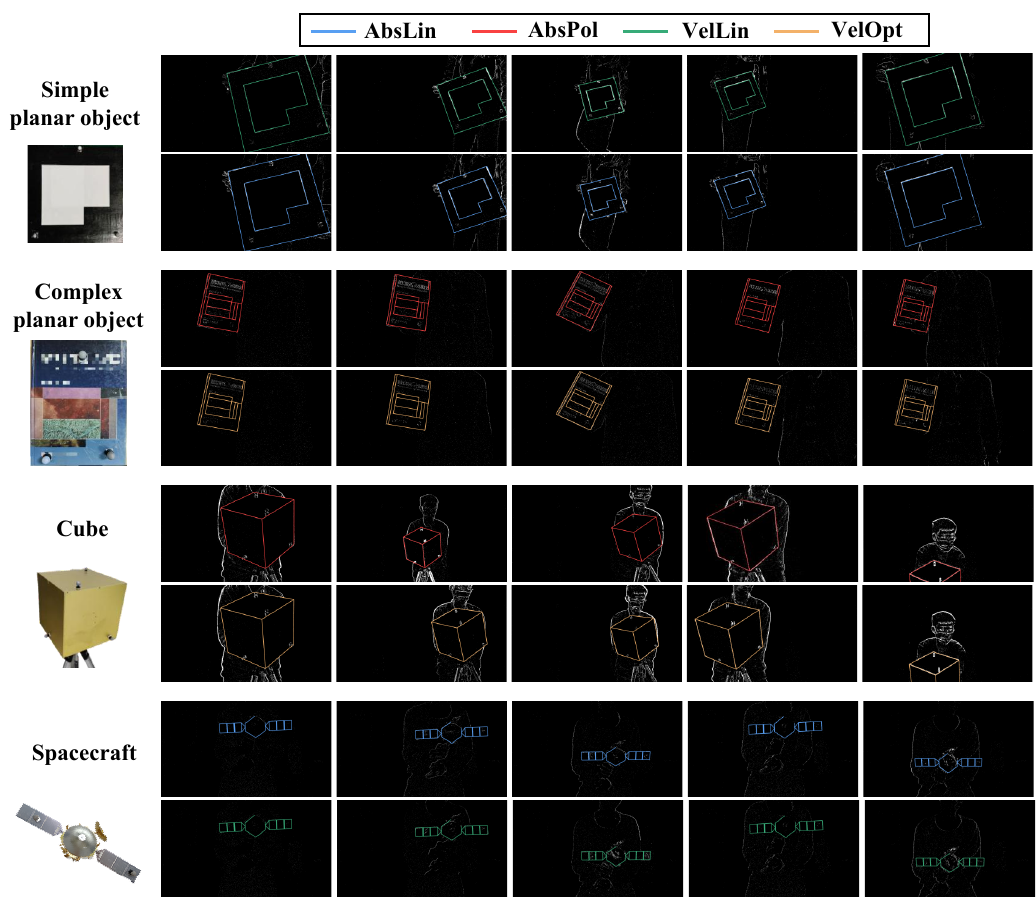}
	\caption{Reprojection results on the LOPET dataset. 3D lines of Objects are reprojected onto the event accumulation images (for visualization purposes only) using the estimated poses.}
	\label{fig_lopet}
\end{figure*}

\paragraph{Experimental results on LOPET}

The LOPET dataset includes objects with diverse linear structures, from simple planar patterns to non-coplanar objects. We evaluate our methods on both initial pose estimation and pose tracking.

We first assess the accuracy of initial pose estimation, a critical step that initializes the entire tracking pipeline. We compare \scenario{AbsLin} and \scenario{AbsPol} with classical approaches like \scenario{OPnP}~\cite{zheng2013revisiting} and \scenario{ASpnL}~\cite{7494617}, as well as the initialization of \scenario{LOPET}~\cite{liu2024line}. Fig.~\ref{fig_initial} presents both a qualitative and quantitative comparison. The reprojection results in Fig.~\ref{fig_initial}\subref{re_vis} visually demonstrate that our method’s estimated pose aligns most precisely with the events. This observation is quantified in Fig.~\ref{fig_initial}\subref{re_qua}, which shows the mean reprojection error of line endpoints. \scenario{AbsPol} achieves the lowest error, confirming its superior accuracy for pose estimation.

We then evaluate the pose tracking performance on the LOPET dataset. The ATE results are summarized in Table~\ref{tab_lopet}. Once again, our proposed methods, \scenario{AbsPol} and \scenario{VelOpt}, deliver state-of-the-art performance. AbsPol achieves the best accuracy on the Complex planar object and Spacecraft sequences. \scenario{VelOpt} excels on the simple planar object. Notably, our method achieves better overall performance in the cube sequence. This sequence is particularly demanding due to partial occlusions, a condition that poses a significant challenge for existing methods, such as \scenario{LS-Opt} and \scenario{EL-SLAM}. The robustness demonstrated by our method under these adverse conditions validates its efficacy and practical advantages. Pose tracking results are shown in Fig.~\ref{fig_lopet}. Across all sequences, the reprojected 3D lines from our estimated poses maintain a precise alignment with the object’s contours in event accumulation images. This demonstrates the stability and accuracy of our tracking methods over extended periods.

\section{Conclusion}
\label{sec:conclusion}
This paper presents an event-driven, geometrically motivated framework for absolute pose and velocity estimation, leveraging 3D lines and their triggered events. Our method leverages two key geometric relations: the orthogonality between event normal vectors and 3D lines, and the collinearity between projected lines and events. These geometric constraints are first utilized to recover absolute pose, and then employed to estimate angular and linear velocities. We propose two types of solvers for pose estimation, namely a linear solver and a polynomial solver. For velocity estimation, we further develop a linear solver and an optimization solver. Extensive experiments on both simulated and real-world datasets demonstrate that the proposed framework achieves strong robustness and accuracy, and consistently outperforms existing methods in terms of accuracy.

\textbf{Limitations and Future Work.} Our method has two main limitations. First, as a geometric solver that serves as a back-end module, it relies on pre-existing 3D line models and robust event–line correspondences, which limits its applicability in map-free scenarios. Second, the method depends on the presence of lines in the scene, which may lead to degraded performance in unstructured environments or areas with sparse texture. To address these limitations, future work will pursue two directions: (1) integrating our solvers with event-based line detectors and event-line SLAM systems to enable joint online line map construction and motion estimation in map-free scenarios; and (2) extending the framework to accommodate hybrid features by jointly utilizing both feature points and lines for more robust event-based motion estimation.

\bibliographystyle{IEEEtran}
\bibliography{refe}

\end{document}